%% file: main.tex
\documentclass[10pt,twocolumn,letterpaper]{article}

\usepackage{cvpr}              

\input{configuration/config}

\usepackage[accsupp]{axessibility}
\usepackage{array,multirow,graphicx}

\def\paperID{5803} 
\def\confName{CVPR}
\def\confYear{2025}

\title{Reconstructing People, Places, and Cameras}

\author{
   Lea M{\"u}ller\textsuperscript{*} \quad
   Hongsuk Choi\textsuperscript{*} \quad
   Anthony Zhang\textsuperscript{} \quad
   Brent Yi\textsuperscript{} \quad
   Jitendra Malik\textsuperscript{}\quad
   Angjoo Kanazawa\textsuperscript{} \\
   \textsuperscript{}UC Berkeley \\
   {\tt\small \{mueller,hongsuk,anthony\_zhang1234,brentyi,malik,kanazawa\}@berkeley.edu} \\
   {\small * equal contribution } \\
}

\begin{document}

\input{sections/02_Content_CameraReady}    
{
    \small
    \bibliographystyle{ieeenat_fullname}
    \bibliography{main}
}

\clearpage
\setcounter{page}{1}
\maketitlesupplementary
\renewcommand{\thetable}{S.\arabic{table}}
\setcounter{table}{0}  
\renewcommand{\thesection}{S.\arabic{section}}
\setcounter{section}{0}  
\renewcommand{\thefigure}{S.\arabic{figure}}
\setcounter{figure}{0}  
\input{sections/04_Appendix_CameraReady}

\end{document}

%% file: configuration/config.tex
\input{configuration/01_Packages}
\input{configuration/02_Highlight}
\input{configuration/03_Style}
\input{configuration/05_Params}

\input{configuration/common_acronyms/bodymodels}




\input{configuration/common_acronyms/metrics}

\input{configuration/common_acronyms/common}

%% file: configuration/01_Packages.tex
\usepackage[accsupp]{axessibility}  
\usepackage[hyphens]{url}
\usepackage[dvipsnames]{xcolor}  
\usepackage{colortbl}
\usepackage{graphicx}
\usepackage{amsmath}
\usepackage{amssymb}
\usepackage{booktabs}
\usepackage{multirow}
\usepackage{cuted}
\usepackage{bm}
\usepackage{xspace}
\usepackage{caption}
\usepackage{balance}
\usepackage{pifont}
\usepackage[normalem]{ulem}
\usepackage{arydshln}
\usepackage{lipsum}
\usepackage{longtable}
\usepackage[utf8]{inputenc}
\usepackage{rotating}
\usepackage{csquotes}
\usepackage{epigraph} 

\usepackage[accsupp]{axessibility}  
\usepackage{colortbl}
\usepackage{graphicx}
\usepackage{amsmath}
\usepackage{amssymb}
\usepackage{booktabs}
\usepackage{multirow}
\usepackage{cuted}
\usepackage{bm}
\usepackage{xspace}
\usepackage{caption}
\usepackage{balance}
\usepackage{pifont}
\usepackage[normalem]{ulem}
\usepackage{arydshln}
\usepackage{lipsum}
\usepackage{times}
\usepackage{epsfig}
\usepackage{graphicx}
\usepackage{amsmath}
\usepackage{amssymb}
\usepackage{tabularx}
\usepackage[hyphens]{url}
\usepackage[dvipsnames]{xcolor}  
\definecolor{cvprblue}{rgb}{0.21,0.49,0.74}
\usepackage{listings}
\usepackage[pagebackref=true,breaklinks=true,colorlinks,bookmarks=false,citecolor=cvprblue]{hyperref}

%% file: configuration/02_Highlight.tex
\newcommand{\ours}{\mbox{HSfM}\xspace}

\newcommand{\cmark}{\textcolor{green}{\ding{51}}} 
\newcommand{\xmark}{\textcolor{red}{\ding{55}}} 

%% file: configuration/03_Style.tex
\newcommand{\colorRef}[1]{\textcolor{black}{#1}} 
\usepackage[capitalize]{cleveref}
\crefname{figure}{\colorRef{Fig.}}{\colorRef{Figs.}}
\Crefname{figure}{\colorRef{Figure}}{\colorRef{Figures}}
\crefname{section}{\colorRef{Sec.}}{\colorRef{Secs.}}
\Crefname{section}{\colorRef{Section}}{\colorRef{Sections}}
\Crefname{table}{\colorRef{Table}}{\colorRef{Tables}}
\crefname{table}{\colorRef{Tab.}}{\colorRef{Tabs.}}

\definecolor{codegreen}{rgb}{0,0.6,0}
\definecolor{codegray}{rgb}{0.5,0.5,0.5}
\definecolor{codepurple}{rgb}{0.58,0,0.82}
\definecolor{backcolour}{rgb}{0.95,0.95,0.92}

\lstdefinestyle{mystyle}{
    commentstyle=\color{codegreen},
    keywordstyle=\color{magenta},
    numberstyle=\tiny\color{codegray},
    stringstyle=\color{codepurple},
    basicstyle=\ttfamily\footnotesize,
    breakatwhitespace=false,         
    captionpos=b,                    
    keepspaces=true,                 
    numbers=left,                    
    numbersep=2pt,                  
    showspaces=false,                
    showstringspaces=false,
    showtabs=false,                  
    tabsize=4,
    morekeywords={with},
    otherkeywords={with},
}

%% file: configuration/05_Params.tex
\newcommand{\energy}{L} 



%% file: configuration/common_acronyms/bodymodels.tex
\newcommand{\smplx}{\mbox{SMPL-X}\xspace}

%% file: configuration/common_acronyms/metrics.tex
\DeclareSymbolFont{matha}{OML}{txmi}{m}{it}
\DeclareMathSymbol{\varv}{\mathord}{matha}{118}

%% file: configuration/common_acronyms/common.tex
\newcommand{\supmat}{Sup.~Mat.\/\xspace}

\newcommand{\twod}{2D\xspace}
\newcommand{\threed}{3D\xspace}

%% file: sections/02_Content_CameraReady.tex
\twocolumn[{
    \renewcommand\twocolumn[1][]{#1}
    \maketitle
    \centering
    \includegraphics[width=1.0 \linewidth,clip,trim=0cm 0cm 0cm 0cm]{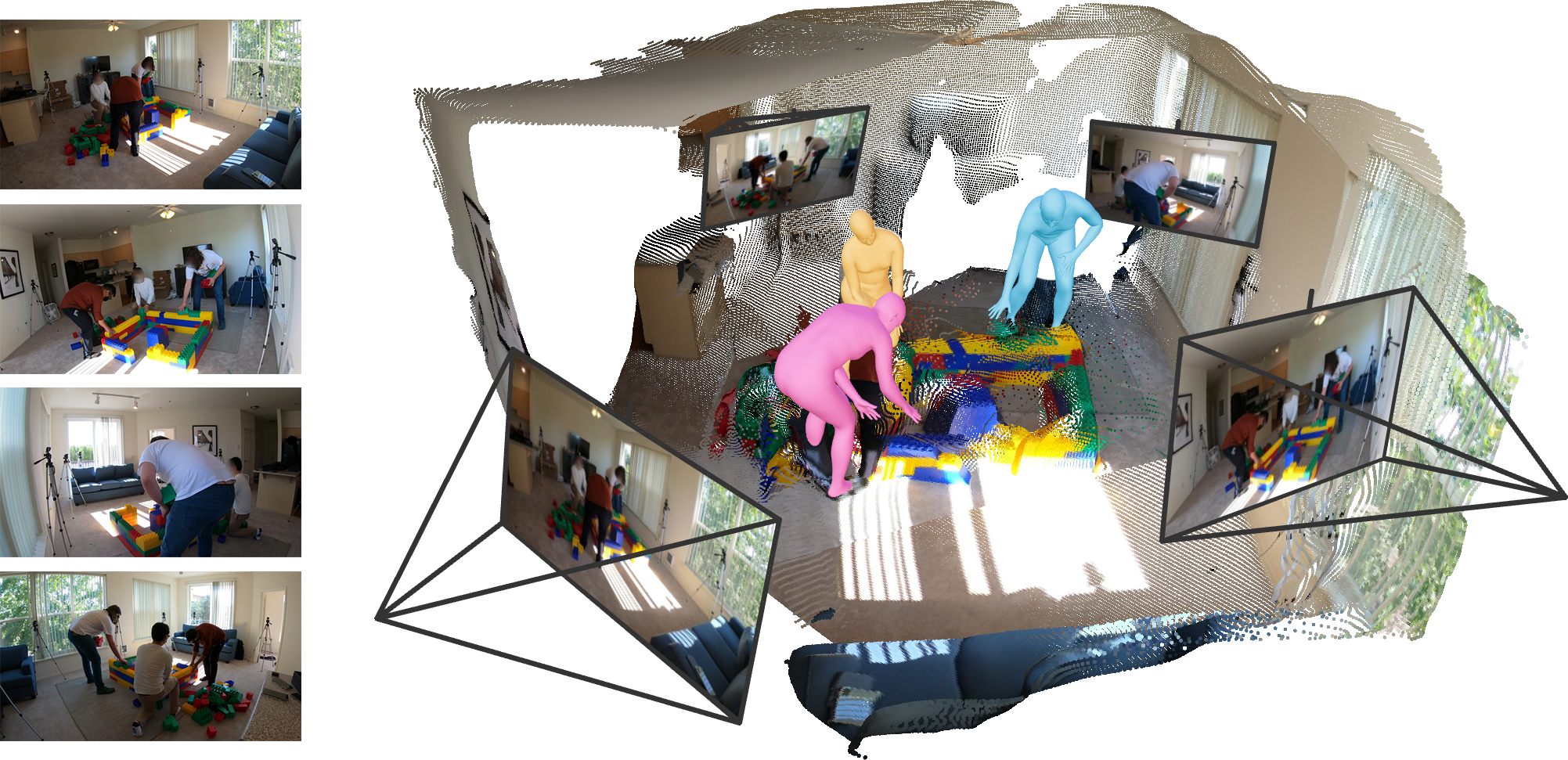}
    \captionof{figure}{\textbf{Humans and Structure from Motion (HSfM).} 
    We propose a method for the joint reconstruction of humans, scene point clouds, and cameras from an uncalibrated, sparse set of images depicting people. By explicitly incorporating humans into the traditional Structure from Motion (SfM) framework through 2D human keypoint correspondences and leveraging robust initialization from an off-the-shelf model for scene and camera reconstruction, our approach demonstrates that integrating these three elements—people, scenes, and cameras—synergistically improves the reconstruction accuracy of each component. Unlike prior work in SfM and human pose estimation, our method reconstructs metric-scale scene point clouds and camera parameters, informed by human mesh predictions, while situating human meshes in coherent world coordinates consistent with the surrounding environment without any explicit contact constraints.
    }
    \vspace{0.9cm}
    \label{fig:teaser_mvmpw} 
}]

\begin{abstract} 
We present ``Humans and Structure from Motion'' (HSfM), a method for jointly reconstructing multiple human meshes, scene point clouds, and camera parameters in a metric world coordinate system from a sparse set of uncalibrated multi-view images featuring people. Our approach combines data-driven scene reconstruction with the traditional Structure-from-Motion (SfM) framework to achieve more accurate scene reconstruction and camera estimation while simultaneously recovering human meshes.
In contrast to existing scene reconstruction and SfM methods that lack metric scale information, our method estimates approximate metric scale by leveraging the human statistical model. Furthermore, our method reconstructs multiple human meshes within the same world coordinate system with the scene point cloud, effectively capturing spatial relationships among individuals and their positions in the environment.
We initialize the reconstruction of humans, scenes, and cameras using robust foundational models and jointly optimize these elements. This joint optimization synergistically improves the accuracy of each component. We compare our method with existing methods on two challenging benchmarks, EgoHumans and EgoExo4D, demonstrating significant improvements in human localization accuracy within the world coordinate frame (reducing error from 3.59m to 1.04m in EgoHumans and from 3.01m to 0.50m in EgoExo4D). Notably, our results show that incorporating human data into the SfM pipeline improves camera pose estimation (\eg, increasing RRA@15 by 20.3\% on EgoHumans). Additionally, qualitative results show that our approach improves scene reconstruction quality. Our code is available at \url{muelea.github.io/hsfm}.
\end{abstract}

\section{Introduction} 
\label{sec:introduction}

In recent years, combining deep learning with multi-view geometry has led to significant advances in two key areas: 3D human reconstruction~\cite{Kanazawa2018_hmr,goel2023humans} and scene reconstruction~\cite{tomasi1992shape,wang2024dust3r}.
However, progress in these domains has largely evolved independently. Human reconstructions often lack anchoring within their surrounding scenes, while scene reconstructions typically exclude people and fail to recover metric scale. In this paper, we propose a unified framework that bridges these two elements.

\begin{figure*}[t]
    \centering
    \includegraphics[width=\linewidth,clip,trim=0cm 0cm 0cm 0cm]{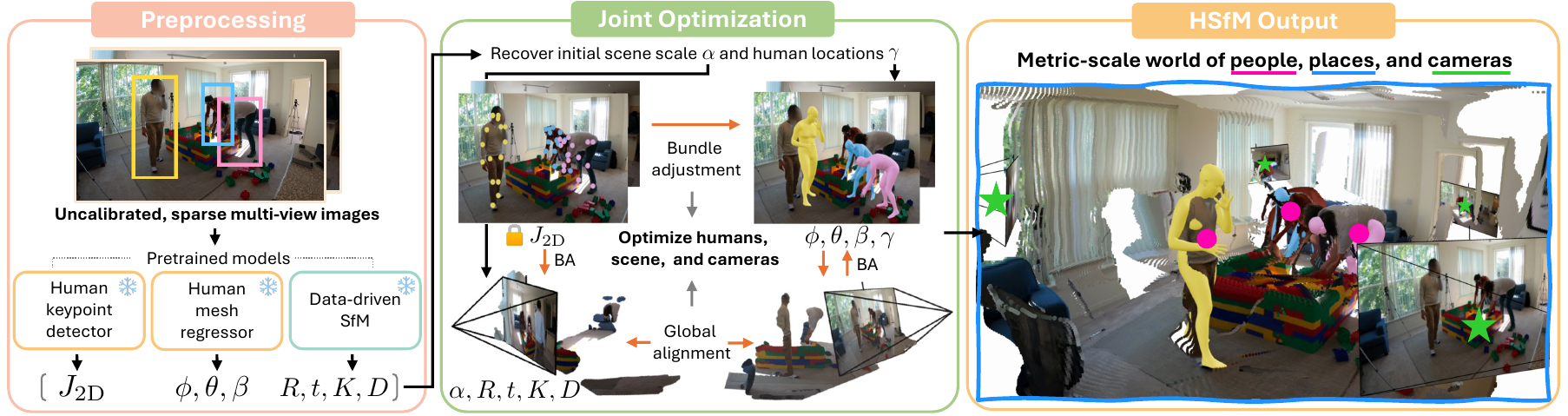}
    \caption[]{\textbf{Pipeline of Humans and Structure from Motion.} Our method processes synchronized images from an uncalibrated multi-view camera setup with known person correspondences across views. We utilize pretrained networks to estimate 2D human keypoints per image~\cite{xu2022vitpose}, 3D human mesh~\cite{goel2023humans}, scene point clouds in a \textit{pointmap} representation, and camera intrinsic and extrinsic parameters~\cite{wang2024dust3r}.
    We first initialize these estimates in a common world coordinate system by recovering the scene scale $\alpha$ and human locations (global translation in the world coordinate) $\gamma$, as described in Section~\ref{subsection:scale_init}. We then jointly optimize humans, the scene, and cameras using bundle adjustment based on 2D human keypoints, 3D human meshes, and a global alignment loss that merges per-view pointmaps into the same world space.
    }
    \label{fig:method}
\end{figure*}

We introduce Humans and Structure from Motion (HSfM), a new method that enables the joint reconstruction of multiple human meshes, scene point clouds, and camera parameters within the same metric world coordinate system as shown in Figure~\ref{fig:teaser_mvmpw}. 
From a sparse set of uncalibrated multi-view images featuring people, our approach combines data-driven scene reconstruction with the traditional Structure-from-Motion (SfM) framework to enhance the accuracy of scene and camera reconstruction while simultaneously estimating human meshes.
The reconstruction process is initialized using robust foundational models for scene reconstruction~\cite{wang2024dust3r} and human reconstruction~\cite{goel2023humans} and further refined through joint optimization. This optimization incorporates a global alignment loss on scene pointmaps and bundle adjustment based on 2D human keypoint predictions~\cite{xu2022vitpose}, significantly enhancing the accuracy of the three components of world reconstruction—humans, scenes, and cameras. 
Our overall pipeline is depicted in Figure~\ref{fig:method}.

Unlike existing approaches to multi-view scene reconstruction~\cite{schonberger2016sfm,schonberger2016pixelwise,yao2018mvsnet,wang2024dust3r} and human pose estimation~\cite{Kanazawa2018_hmr,choi2020pose2mesh,xu2022multi}, HSfM recovers the metric scale of scene point clouds and camera poses while situating human meshes within a unified world coordinate system.
The comprehensive output of HSfM facilitates the capture and evaluation of spatial relationships among individuals, ensuring consistency with the surrounding environment. Furthermore, unlike prior multi-view human pose estimation methods that depend on precise camera calibration~\cite{dong2019fast,huang2022rich,yuan2022humman}, our approach operates with minimal constraints on the capture setup and does not require prior knowledge of the environment.

Our approach is founded on two key insights. The first insight is that deep learning-based human mesh estimation inherently contains metric scale information, as the predictions reflect the statistical human size present in the training datasets, thereby constraining the scale of the scene. The second insight is that robust 2D human keypoint predictions and 3D human mesh estimations provide precise correspondences and reliable initial 3D structures for bundle adjustment. Note that for the purpose of this work, we assume known re-identification of people across camera views.

We evaluate our approach on two challenging benchmarks, EgoHumans~\cite{khirodkar2023egohumans} and EgoExo4D~\cite{grauman2024ego}, which feature individuals participating in a variety of indoor and outdoor activities across diverse environments.
We assess the accuracy of human mesh reconstruction by comparing our method to other approaches that estimate human poses in a world coordinate frame~\cite{xu2022multiview}. Additionally, we compare camera pose accuracy against learning-based dense scene reconstruction methods, such as DUSt3R~\cite{wang2024dust3r} and MASt3R~\cite{leroy2024mast3r}.
Our approach demonstrates substantial improvements in camera pose estimation compared to existing methods while accurately positioning individuals within the scene. Specifically, it achieves approximately a 3.5-fold improvement in human metrics, reducing the human world location error from 3.51m to 1.04m on EgoHumans, and delivers camera metric improvements of approximately 2.5 times compared to the most relevant baseline~\cite{xu2022multi}. These results underscore the effectiveness of our method, which leverages the joint reconstruction of multiple human meshes, scene point clouds, and cameras, supported by robust initialization for humans~\cite{goel2023humans} and cameras~\cite{wang2024dust3r}.
We further validate our design through ablations which show the synergy between humans, scenes, and cameras. 
Our qualitative results highlight that the joint optimization of people (multiple humans), places (scenes), and cameras not only enhances human localization but also improves scene reconstruction and camera pose estimation.

In summary, we present Humans and Structure from Motion (HSfM), which provides a comprehensive representation of the world—encompassing people, places, and cameras—marking a step forward in understanding complex real world environments.

\section{Related Work} 
\label{sec:relatedwork}

\begin{table}[t]
    \centering
        \resizebox{\columnwidth}{!}{
            \begin{tabular}{lcccccc}
                \toprule
                & \textbf{Local Pose}  & \textbf{Multi-Person} & \textbf{Stat. Scale} & \textbf{Camera} & \textbf{Places} \\
                \midrule
                HMR2 \cite{Kanazawa2018_hmr,goel2023humans}    & \cmark & \xmark & \xmark & \xmark & \xmark \\
                Multi-HMR \cite{baradel2024multi} & \cmark & \cmark & \cmark & \xmark & \xmark  \\
                SLAHMR \cite{ye2023decoupling}  & \cmark  & \cmark & \cmark & \cmark & \xmark  \\
                UnCaliPose \cite{xu2022multi}  & \cmark & \cmark & \cmark & \cmark & \xmark  \\
                DUSt3R \cite{wang2024dust3r} & \xmark & \xmark & \xmark & \cmark & \cmark \\
                MASt3R \cite{leroy2024mast3r} & \xmark  & \xmark & \cmark & \cmark & \cmark \\
                \midrule
                \textbf{\ours}  & \cmark  & \cmark & \cmark & \cmark & \cmark \\
                \bottomrule
            \end{tabular}
            }
    \caption{\textbf{Comparison of methods across different features}. Previous works in human pose estimation like HMR2, SLAHMR, Multi-HMR, and UnCaliPose has made great progress in reconstructing body poses in single- and multi-person setups from images. Recent methods like DUSt3R and MASt3R are able to recover accurate camera poses and scene points clouds (places). This includes reconstructions with statistically correct scale (Stat. Scale) which can be obtained from human body models, \eg in SLAHMR, or, as in MASt3R, from world knowledge. Our approach, \ours, is the first  to jointly reconstruct multiple people, scene, and cameras from sparse uncalibrated multi-view images.
    }
    \label{tab:comparison}
\end{table}

Research in multi-person mesh reconstruction and Structure from Motion has seen substantial progress showing remarkable domain-specific results (see \cref{tab:comparison}). Building on these foundational works, our approach unifies these areas.

\textbf{Structure from Motion.}
Structure from Motion (SfM)~\cite{beardsley1996model,fitzgibbon1998automatic,hartley2004multiple} aims to reconstruct camera poses and 3D scene geometry from a set of images by establishing pixel correspondences across views. 
Traditional SfM Methods~\cite{agarwal2011building,crandall2013sfm,cui2017hsfm}, such as COLMAP~\cite{schonberger2016sfm,schonberger2016pixelwise} employs keypoint detection, matching based on locally invariant descriptors~\cite{lowe2004distinctive,bay2006surf,rublee2011orb}, and incremental bundle adjustment to estimate camera poses and sparse 3D points.
However, these traditional approaches are highly sensitive to noise at each stage of their sequential pipeline and require specific conditions for input, such as a large number of camera views with substantial overlapping image areas.

Learning-based SfM methods replace one or more components of the traditional SfM pipeline with data-driven approaches~\cite{yi2016lift,detone2018superpoint,barroso2019keynet,revaud2019r2d2,lindenberger2023lightglue}. Recently, dense matching-based SfM~\cite{truong2020glunet,jiang2021cotr,sun2021loftr,truong2021densecorrespondences,chen2022aspanformer,zhu2023pmatch} has shifted from sparse keypoints to dense, data-driven approaches.
DUSt3R~\cite{wang2024dust3r} and more recent works \cite{duisterhof2024mast3r_sfm, leroy2024mast3r} exemplify this by predicting dense 3D pointmaps without requiring camera calibration.
However, as these methods primarily focus on scene structure, they do not estimate human poses and face challenges in reconstructing pixels corresponding to people. In contrast, our approach robustly recovers both human poses and scene structure simultaneously.

\textbf{SfM with Humans.} Several recent works have leveraged humans in the scene as cues to overcome the limitations of traditional Structure from Motion (SfM) methods in challenging scenarios with minimal overlap or wide baselines. Ma et al. \cite{ma2022virtual,pavlakos2022human} introduce the concept of Virtual Correspondences (VCs), which are pairs of pixels from different images whose camera rays intersect in 3D space, even if the points are not co-visible. 
Similarly, Xu et al. \cite{xu2021widebaseline} addressed wide-baseline multi-camera calibration by employing 2D keypoint associations from people across different cameras, obtained by person re-identification methods. Xu and Kitani~\cite{xu2022multiview} extended this work by sequentially solving for person re-identification, camera pose estimation, and 3D human pose estimation using multi-view geometry and bundle adjustment in an optimization pipeline. While our approach aligns with this line of research, it differs by jointly optimizing people, places (scene pointmaps), and cameras.

\textbf{Multi-view human reconstruction.} In controlled environments with known camera parameters, multi-view reconstruction leverages geometric consistency for accurate 3D pose estimation, reducing single- and multi-person tasks to a triangulation problem~\cite{triggs2000bundle} with a long history of research. Recent works explore setups with unknown camera poses by employing end-to-end learning methods that jointly estimate camera parameters and 3D poses~\cite{yu2022multiview, hardy2023unsupervised}. However, these methods are often limited to single-person scenarios~\cite{yu2022multiview} or do not incorporate scene context~\cite{hardy2023unsupervised}. Existing multi-person methods focus on re-identification~\cite{chen2020multiperson,dong2019fast,kadkhodamohammadi2018generalizable} or, for video, on re-identification and tracking~\cite{huang2022rich}. In contrast to previous work, our approach does not require camera calibration. Instead, we leverage the human body structure and data-driven SfM methods to achieve accurate human pose and camera estimates.

\section{Preliminaries and Notation}
\label{sec:notation}

\noindent \textbf{Setup.} Our method takes as input an uncalibrated, sparse set of $C$ images capturing people in a scene at a single moment in time. We denote each image as $I^{c}$, $c = \{1 \dots C\}$, corresponding to each camera, with resolution $H^c \times W^c$. We assume humans have been associated across views. Given this input, our method jointly reconstructs humans, scene, and cameras in a metric 3D world.

\noindent \textbf{Human.} For all of the following, we represent humans in the scene via a human body model, SMPL-X \cite{Pavlakos2019_smplifyx}. SMPL-X is a differentiable function that maps pose, $\theta \in \mathrm{SO}(3)^{J}$, and shape, $\beta \in \mathbb{R}^B$ to a triangulated mesh with $J$ joints. This mesh can be placed in the world via two additional parameters, orientation, $\phi \in \mathrm{SO}(3)$, and translation, $\gamma \in \mathbb{R}^3$. We model multiple people, \ie $h \in \{1 \dots H\}$ humans.
In summary, a human, $h$, in the world is defined via 
\begin{equation}
\begin{split}
H^h = \{\phi^h, \theta^h, \beta^h, \gamma^h \} \text{.}
\end{split}
\end{equation}

\noindent \textbf{Cameras.} To project 3D points onto an image, $I \in \mathbb{R}^{H \times W \times 3}$, we use a perspective camera model with intrinsics, $K \in \mathbb{R}^{3 \times 3}$ with focal lengths $(f_x, f_y)$ and principal point $(W/2, H/2)$, and extrinsics with rotation, $R \in \mathrm{SO}(3)$, and translation, $t \in \mathbb{R}^3$. Existing methods produce camera estimates that are not necessarily scaled to real-world size. To address this, we introduce a scaling parameter, $\alpha$, which adjusts the distance between cameras while preserving their relative directions. 
With these parameters a 3D point, $x^{\text{3D}}$
can be projected to 2D via 
\begin{equation}
\begin{split}
x_{\text{2D}} = K (R x_{\text{3D}} + \alpha t) \text{.}
\end{split}
\end{equation}
The pixel coordinates, $(u', v') = (\frac{u}{w}, \frac{v}{w})$ are obtained from $x_{\text{2D}} = [u,v,w]^\top$. $K^c$ and $R^c / t^c$ denote the in- and extrinsics of camera $c \in \{1, \dots C\}$.

\noindent \textbf{Scene.} We represent the scene via per-view pointmaps \cite{wang2024dust3r}, $\mathcal{S} \in \mathbb{R}^{W \times H \times 3}$, a dense  pixel-aligned 3D location for its corresponding image $I$ in the world coordinate frame. 
$S^c$ denotes the pointmap of an image $c$. A nice property of pointmap formulations is that we can express them through camera estimates and depth maps. For an image pixel $(i,j)$ , its corresponding pointmap's world coordinate can be written as
\begin{equation}
\begin{split}
\mathcal{S}_{i,j} = \alpha (R^\top [K^{-1}  D_{i,j}[i,j,1]^\top] - R^\top t) \text{.}
\end{split}
\end{equation}
This formulation unprojects a pixel $(i,j)$ using its depth value $D_{i,j}$ and $K$, and maps it to the world coordinate system through $R^\top$, $-R^\top t$, \ie the \textit{camera-to-world} transformation defined by $R$ and $t$, and scaling.

\section{Humans and Structure from Motion}
\label{sec:reconstruction}
Our method takes as input an uncalibrated, sparse set of images capturing people in a scene at a single moment in time. Given this input, our goal is to jointly estimate each person’s human parameters, the scene, and the camera parameters.
Our key insight is that jointly reasoning about people, scene structure, and cameras improves all three aspects of reconstruction. To achieve this, we integrate global scene optimization from recent scene reconstruction methods with the traditional Structure-from-Motion (SfM) formulation. This integration leverages 2D human keypoints as reliable correspondences and 3D human meshes as robust 3D structures for bundle adjustment. Please also refer to Figure~\ref{fig:method}.

Our joint optimization approach has several advantages. By incorporating human mesh predictions, the method introduces metric scale into the reconstruction process, leveraging statistical information about human body dimensions. The cameras and 2D human keypoints enable precise positioning of individuals within the world coordinate system, allowing for the recovery of their heights and relative distances. Additionally, correspondences of 2D human keypoints enhance camera calibration, which in turn improves scene reconstruction. The scene structure further stabilizes the camera pose registration, creating a feedback loop that refines the overall system. The result is a globally consistent reconstruction of humans, scenes, and cameras, providing a comprehensive understanding of the environment.
Since this is an under-determined problem, we take advantage of data-driven 3D human~\cite{goel2023humans} and scene~\cite{wang2024dust3r} reconstruction methods to provide initializations. Note that our method can easily integrate other mesh regressors tailored towards estimating non-standard body size such as BEV~\cite{sun2022putting}.

\subsection{Initialization of World}
\label{subsection:scale_init}
The initial estimates of humans, scene structure, and cameras are derived from different networks and therefore exist in separate coordinate systems. Our objective is to align these components within a unified world coordinate system, a process we refer to as the initialization of the world. To achieve this, we estimate the metric scale that aligns the scene pointmaps and cameras with the humans. 

One may simply start the optimization by setting the scale $\alpha=1$. 
However, since SfM reconstructions are up-to-scale, the magnitude of $t$ may vary significantly in every reconstruction.
If the SfM scene is too small relative to people, cameras may be in-front of the humans,~\eg when the scene is placed inside the human mesh, leading to degenerate solutions. Setting alpha to arbitrary big value can prevent this, but makes the problem prone to local minima.

We propose an analytical method to approximate a consistent reconstruction using data-driven outputs of 2D/3D human keypoints and camera parameters. Initially, we roughly position individuals within the scene by estimating $\gamma$, based on the predicted 2D/3D human keypoints and the focal length from $\tilde{K}$, following a similar approach to Ye et al.~\cite{ye2023decoupling}. Next, we calculate the initial scale $\alpha$ of the data-driven SfM-predicted world by aligning the data-driven with human-centric camera positions.

Specifically, we first obtain each camera's rotation, $\hat R^c$, using the estimated 3D human body orientation, $\tilde \phi$.
This leverages the fact that the human's orientation should remain consistent in the world coordinate frame across different views. Assuming a reference camera, $c_1$, and an anchor person, $\tilde \phi^h$, we recover each camera's rotation by solving
\begin{equation}
    \begin{split}
(\hat R^c)^\top = (R^{c_1})^\top \tilde \phi^{h c_1} (\tilde \phi^{h c})^\top.
\end{split}
\end{equation}
We pick the anchor person $h$ based on the best view coverage \wrt to the 2D joint confidence scores. 

We estimate the camera translation by first estimating the location of the person $\gamma$ in the world using the data-driven focal length prediction from $\tilde K$ and the size of the predicted human following the similar triangle ratio using the average 2D and 3D bone lengths as done in Ye~\etal~\cite{ye2023decoupling}. 
Because the human position in the world coordinate frame should remain consistent when viewed from any camera, we recover camera position $T^c$ in the world coordinate frame via

\begin{equation}
\begin{split}
\hat T^{c} = \tilde \gamma^{c_1} - (\hat R^c)^\top \tilde \gamma^{c}.
\end{split}
\end{equation}
Finally, given the camera positions $\hat T$ derived from the humans and the data-driven camera position predictions $\tilde T$, we solve a least-squares problem to compute $\hat \alpha$, which aligns the scene pointmap $\mathcal{S}$ and the camera translation $t$ with the metric-scale world defined by the humans $\tilde H$.
The scaling factor $\hat \alpha$ provides a reasonable approximation of the metric scale.
For the initial estimates of $\tilde H$, we rely on an estimate from a reference camera in practice.
Please see the project page supplementary video for an animated illustration. 

\subsection{Reconstructing People, Places, and Cameras}
After initializing the world, we jointly optimize the humans, depth maps, and cameras using a global scene optimization loss and bundle adjustment guided by 2D human keypoint predictions. The objective function is defined as follows:
\begin{equation}
    \min_{\{\alpha, \gamma, \beta, \phi, \theta, R, t, K, D\}} L_{\text{Humans}} + \lambda L_{\text{Places}} \text{.}
\label{eq:objective}
\end{equation}

To gradually guide the optimization towards the global minimum, we first optimize $\{\alpha, \gamma, \beta\}$ with $\lambda = 0$. Then, we set $\lambda$ and optimize $\{\gamma, \beta, \phi, \theta, R, t, K, D\}$. The result of this optimization is a metric-scale world with consistent humans, scene, and cameras.

\noindent\textbf{Bundle adjustment based on human keypoints:} 
We define the bundle adjustment objective as follows:
\begin{equation}
    \begin{split}
        L_{\text{Humans}} = \frac{1}{HC}\sum_{c=1}^{C} \sum_{h=1}^{H} \energy^{ch}_{J} + \frac{1}{H} \sum_{h=1}^{H} & \energy^h_{\beta} \text{.}
    \end{split}
\end{equation}

\noindent The term $\energy_{J}$ denotes the re-projection error between the \threed joints of the current estimate with the estimated camera and detected \twod keypoints by ViTPose~\cite{xu2022vitpose}:
\begin{equation}
    \begin{split}
\energy^{ch}_{J} = \frac{1}{b^{ch}_{\text{2D}}}||c^{ch}_{\text{2D}} (J^{ch}_{\text{2D}} - K^{c}( R^{c} J^{h}_{\text{3D}} + \alpha t^c))||_2
    \end{split}
    \label{eq:reprojection}
\end{equation}
The keypoint loss for each person, $h$, and camera, $c$, is normalized by the bounding box height $b^{ch}_{\text{2D}}$ of the 2D human detection and detected keypoints weighted by their estimated confidence scores $c^{ch}_{\text{2D}}$.  We further regularize human body shape, $\tilde \beta$, to stay close to the average shape of the human body model: 
\begin{equation}
    \begin{split}
        \energy^{h}_{\beta} & = ||\beta^{h}||_2 \text{.} 
    \end{split}
\end{equation}
The output of this optimization is people in plausible size and locations in the world coordinate frame. Please refer to the supplementary material for implementation details.

\noindent\textbf{Global scene optimization:}
We adapt the global alignment loss from DUSt3R~\cite{wang2024dust3r}, which originally optimizes camera and world pointmaps.
Intuitively, this loss takes pairs of cameras, say A and B ($c_i$ and $c_j$), with the predicted content of $c_i$ in B's view. The alignment loss transforms this ``A's-content-in-B's-view'' to the world coordinate frame where it's compared against the optimized global scene pointmap, which is A's content in the world. 

More formally, the global alignment loss aligns per-view pointmaps into a joint world space, \ie $\{ \mathcal{S}^{c} \in \mathbf{R}^{H \times W \times 3} \}$ for $c = 1 \dots C$. To achieve this, the alignment loss takes cross-view pointmaps $X^{c_i, c_j}$ for two cameras $(c_i, c_j) \in \mathcal{E}$, with $i,j = 1 \dots C$ and $\mathcal{E}$ being the set of all pairs of cameras with $i \neq j$. The notation, $X^{c_i, c_j}$, describes cross-view pointmaps, meaning that the pointmap $X^{c_i}$ of camera $c_i$ is expressed in the coordinate frame of camera $c_j$. Pairwise transformation matrices $P^{c_i,c_j \rightarrow w} \in \mathbb{R}^{3 \times 4}$, \ie the transformation matrix for camera pair $c_i$, $c_j$ that brings $c_j$'s content to the world coordinate frame. The global alignment loss term is defined as

\begin{equation}
\small
    L_{\text{Places}}   = 
        \sum_{(c_i, c_j) \in \mathcal{E}} \sum_{i = 1}^{HW} Q_i^{c_i,c_j} || \mathcal{S}_i^{c_i} - \sigma^{c_i,c_j \rightarrow w} P^{c_i,c_j \rightarrow w} X_i^{c_i, c_j} || \text{,}
\end{equation}

\noindent where $Q_i^{c_i,c_j}$ are predicted per-pixel confidence maps. $\sigma^{c_i,c_j \rightarrow w}$ is a scaling factor associated to the pair $(c_i,c_j)$. Note that different from DUSt3R \cite{wang2024dust3r}, we don't regularize it to avoid a trivial optimum, since the scale is constrained by humans.
We omit $\sigma$ and $P$ in Eq.~\ref{eq:objective} for clarity, but they are still optimized together following DUSt3R.

\begin{figure*}
    \centering
    \begin{subfigure}[b]{0.49\linewidth}
        \centering
        \includegraphics[width=\linewidth]{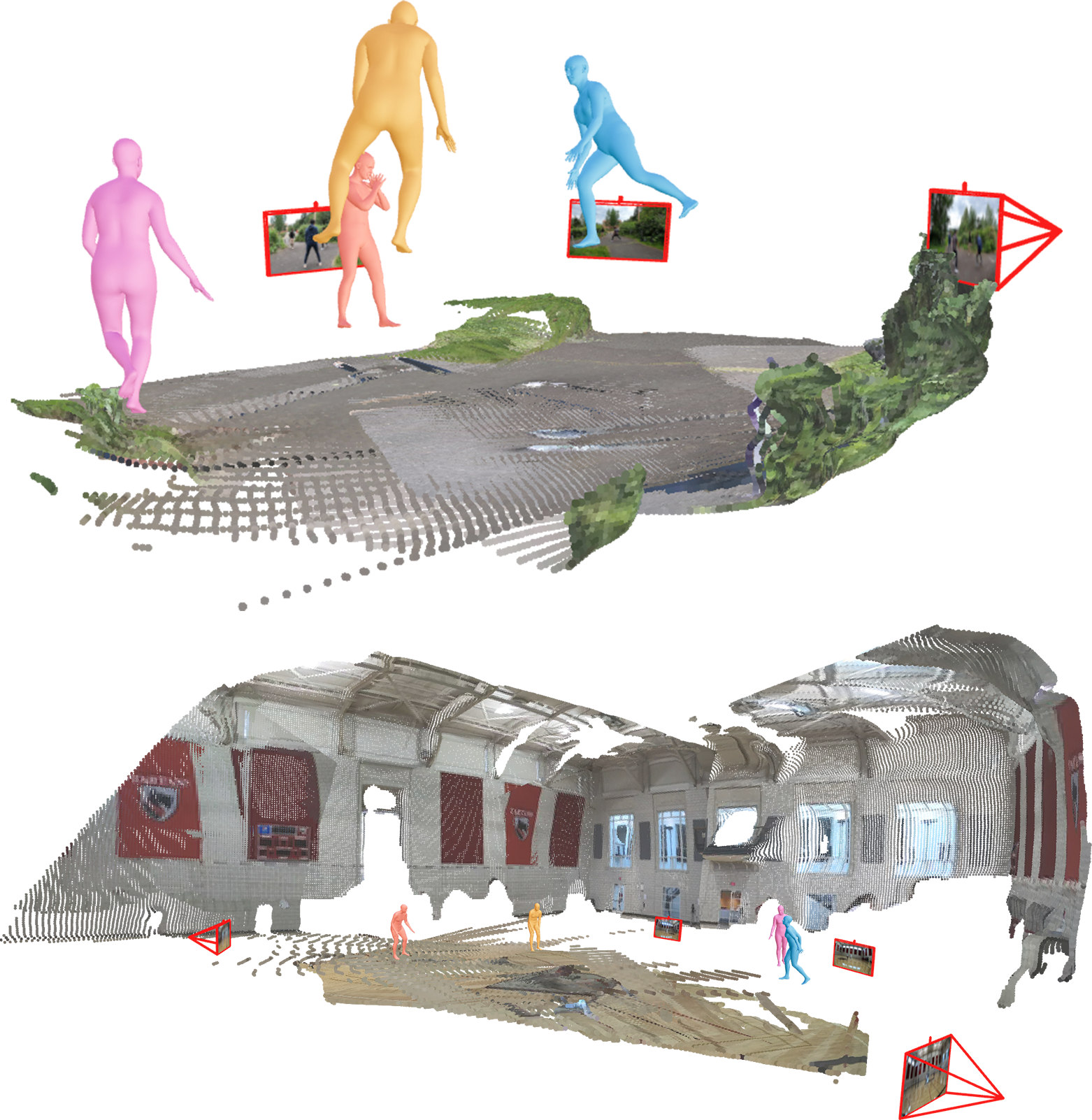}
        \caption{EgoHumans - Initial state}
    \end{subfigure}
    \hfill
    \begin{subfigure}[b]{0.49\linewidth}
        \centering
        \includegraphics[width=\linewidth]{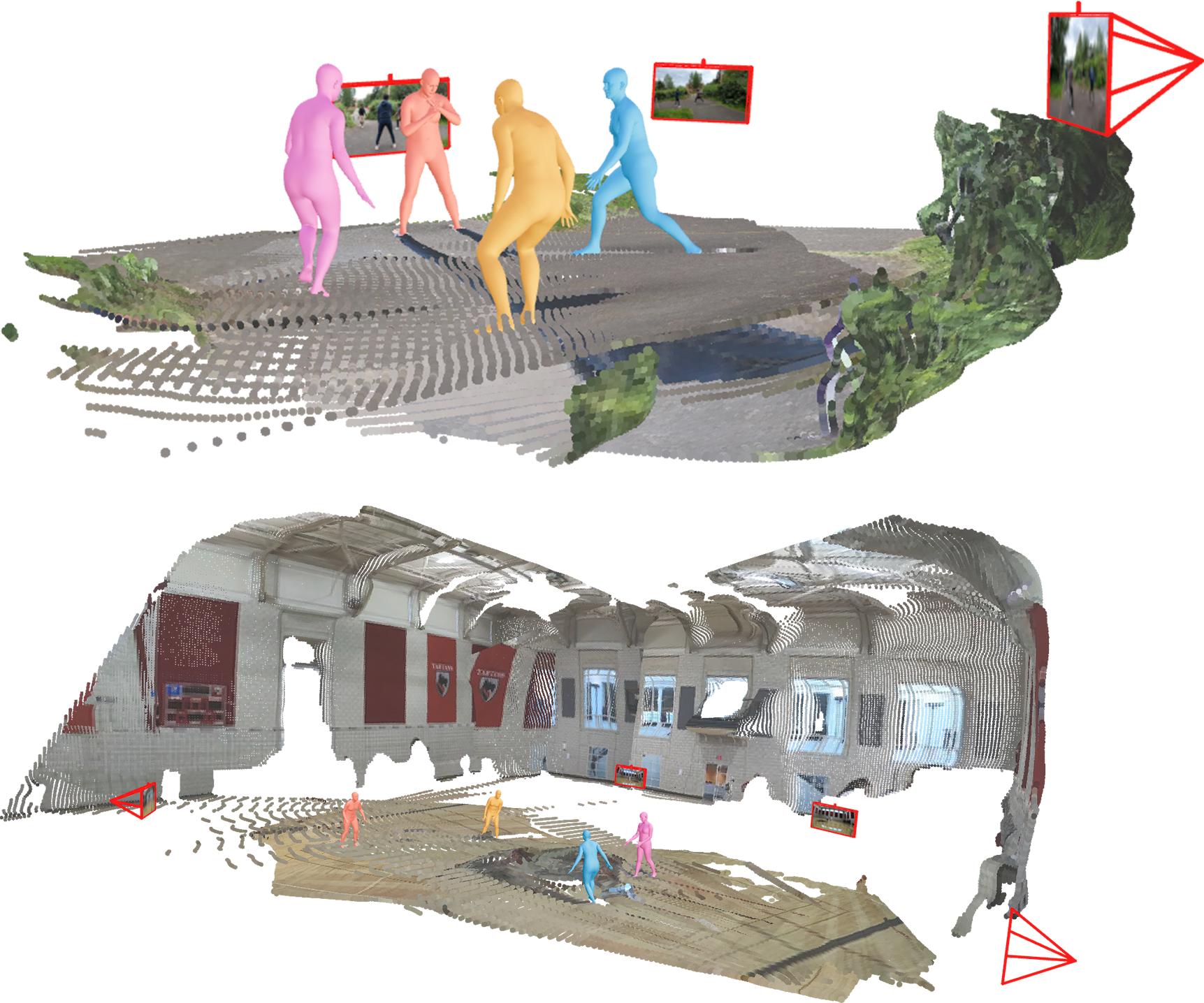}
        \caption{EgoHumans - After optimization}
    \end{subfigure}
    
    \vspace{1em}
    \begin{subfigure}[b]{0.24\linewidth}
        \centering
        \includegraphics[width=\linewidth]{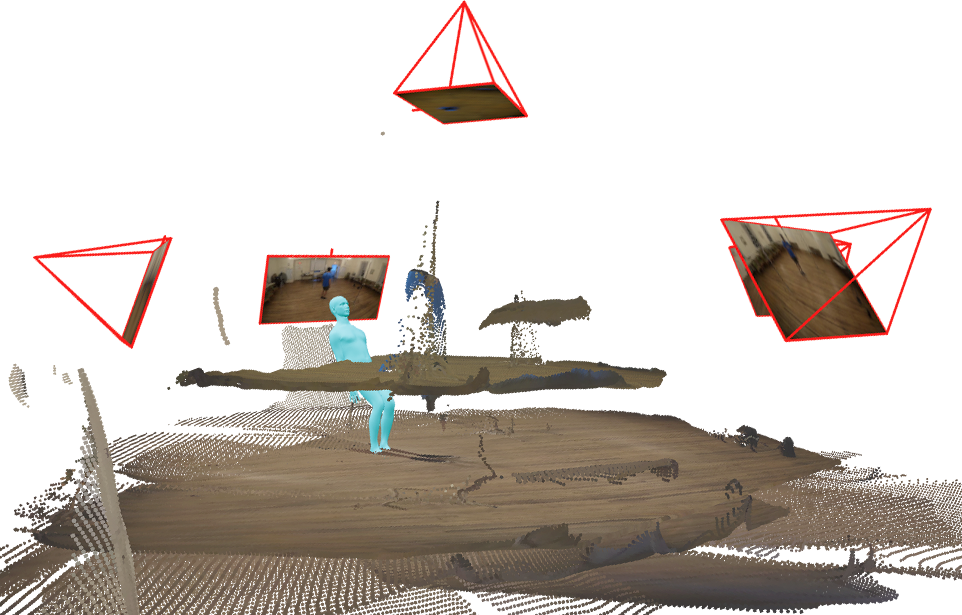}
        \caption{EgoExo Scene 1 - Initial}
    \end{subfigure}
    \hfill
    \begin{subfigure}[b]{0.24\linewidth}
        \centering
        \includegraphics[width=\linewidth]{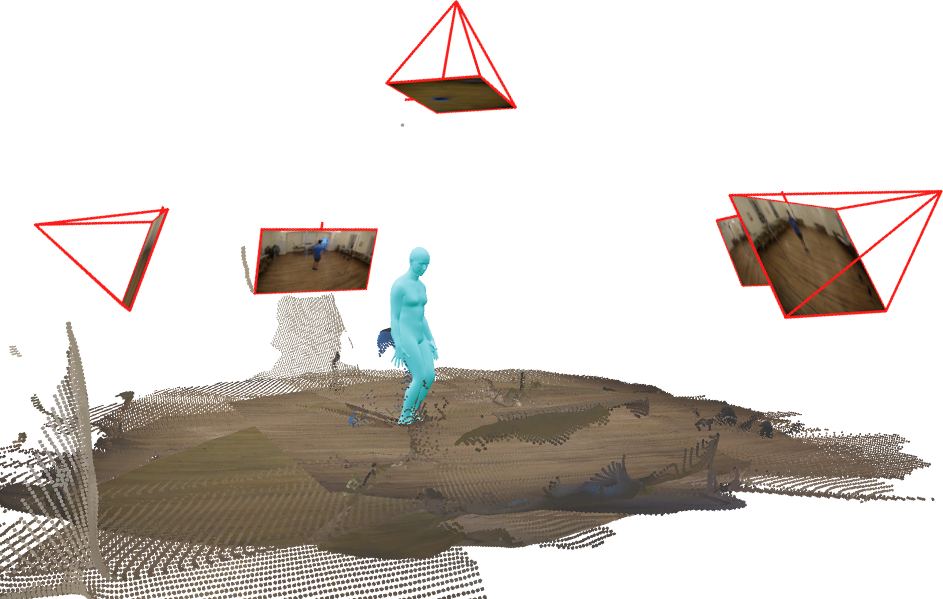}
        \caption{EgoExo Scene 1 - Optimized}
    \end{subfigure}
    \hfill
    \begin{subfigure}[b]{0.24\linewidth}
        \centering
        \includegraphics[width=\linewidth]{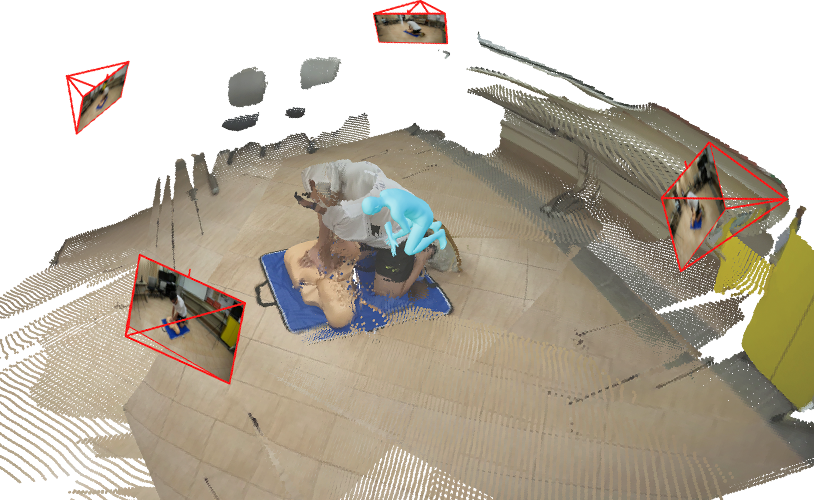}
        \caption{EgoExo Scene 2 - Initial}
    \end{subfigure}
    \hfill
    \begin{subfigure}[b]{0.24\linewidth}
        \centering
        \includegraphics[width=\linewidth]{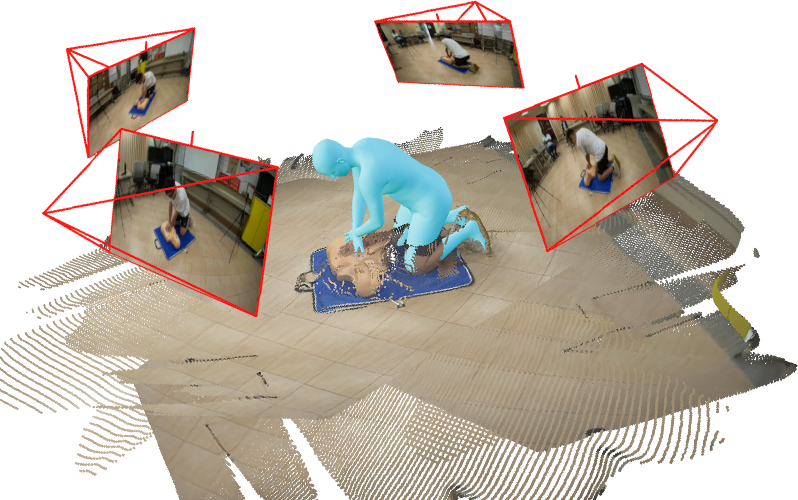}
        \caption{EgoExo Scene 2 - Optimized}
    \end{subfigure}
    \caption{\textbf{Qualitative results from \ours.} We show our optimized result on sequences from EgoHumans (top) and EgoExo4D (bottom). Note how in the Initial state (left) people are floating in the air (a), how the scene and human scale is not aligned (e), and how noisy the scene appears (c). Our method resolves these problems by grounding people in the scene (b), recovering plausible metric scale (f), and better camera estimates (d). We achieve this without scene contact constraints, which often require assumptions about the environment—such as flat terrain—or about motion, such as the assumption that humans are always in contact with the ground (i.e., no jumping). For more qualitative results, including a demo on images taken in the wild with a minimal capturing setup, please see our supplementary material.}
    \label{fig:big}
\end{figure*}

\section{Experiments} 
\label{sec:experiments}

\begin{table*}
    \footnotesize
    \centering
    \setlength{\tabcolsep}{0.9pt}
        \begin{tabular}{clccccccccccccc}
            \toprule
             \multirow{2}{*}{} & \multicolumn{1}{c}{\multirow{2}{*}{Method}} & \multicolumn{3}{c}{\textbf{Human Metrics}} & \multicolumn{10}{c}{\textbf{Camera Metrics}} \\
             & \multicolumn{1}{c}{} & W-MPJPE$\downarrow$ & GA-MPJPE$\downarrow$ & PA-MPJPE$\downarrow$ & TE$\downarrow$ & s-TE$\downarrow$ & AE$\downarrow$ & RRA@10$\uparrow$ & RRA@15$\uparrow$ & CCA@10$\uparrow$ & CCA@15$\uparrow$ & s-CCA@10$\uparrow$ & s-CCA@15$\uparrow$ \\
            \midrule
                        \multirow{5}{*}{\rotatebox[origin=c]{90}{\color{olive} EgoHumans}} & UnCaliPose* \cite{xu2022multi} & 3.51 & 0.67 & 0.13& 2.63& 2.63 & 60.90 & 0.28 & 0.39 & - & - & 0.33 & 0.44 \\
            & DUSt3R \cite{wang2024dust3r} & - & - & - & -  & 1.15 & 11.00 & 0.61 & 0.74 & - & - & 0.49 & 0.74 \\
            & MASt3R \cite{leroy2024mast3r} & - & - & - & 4.97 & 0.92 & 10.42 & 0.61 & 0.74 & 0.06 & 0.07 & 0.65 & 0.86 \\
            & \ours (init.) & 4.28 & 0.51 & 0.06 & 2.37 & 1.15 & 11.00 & 0.52 & 0.79 & 0.26 & 0.38 & 0.49 & 0.74 \\
            & \textbf{\ours (Ours)} & \textbf{1.04} & \textbf{0.21} & \textbf{0.05} & \textbf{2.09} & \textbf{0.75} & \textbf{9.35} & \textbf{0.72} & \textbf{0.89} & \textbf{0.32} & \textbf{0.46} & \textbf{0.75} & \textbf{0.91} \\ \midrule
            \multirow{5}{*}{\rotatebox[origin=c]{90}{\color{olive} EgoExo4D}} & UnCaliPose* \cite{xu2022multi} & 3.59 & - & 1.19 & 2.21 & 0.98 & 63.98 & 0.20 & 0.31 & - & - & 0.26 & 0.37 \\
            & DUSt3R \cite{wang2024dust3r} & - & - & - & -  & \textbf{0.34} & 10.06 & \textbf{0.81} & 0.88 & - & - & 0.64 & \textbf{0.84} \\
            & MASt3R \cite{leroy2024mast3r} & - & - & - & 1.03 & 0.36 & \textbf{9.11} & \textbf{0.81} & \textbf{0.90} & \textbf{0.09} & \textbf{0.17} & \textbf{0.70} & 0.81 \\
            & \ours (init.) & 5.80 & - & 0.08 & 1.27 & \textbf{0.34} & 10.06 & \textbf{0.81} & 0.88 & 0.05 & 0.10 & 0.64 & \textbf{0.84} \\
            & \textbf{\ours (Ours)} & \textbf{0.50} & - & \textbf{0.07} & \textbf{1.01} & \textbf{0.34} & 10.39 & 0.80 & 0.89 & 0.05 & 0.14 & \textbf{0.70} & \textbf{0.84} \\ 
            \bottomrule
        \end{tabular}
    \caption{\textbf{Evaluation on EgoHumans and EgoExo4D.}  
    \ours outperforms existing human and scene reconstruction methods, delivering metric-scale reconstructions for humans, the scene, and cameras within the same world coordinate system. Our approach shows major improvement over the initial estimates \ours (init.), obtained from DUSt3R~\cite{wang2024dust3r} and HMR2~\cite{goel2023humans} (Section~\ref{subsection:scale_init}), particularly when multiple humans are present in the scene, as seen in EgoHumans, compared to a single individual in EgoExo4D.
    Additionally, \ours surpasses MASt3R in metric-scale camera metrics, demonstrating the benefit of human size for recovering scene scale. All baselines and our method use four cameras on EgoHumans and up to six cameras on EgoExo4D. Human and camera translation metrics are reported in meters.
    }
    \label{tab:egohumans}
\end{table*}

\begin{table*}
    \footnotesize
    \centering
    \setlength{\tabcolsep}{0.9pt}
        \begin{tabular}{lccccccccccccc}
            \toprule
             \multicolumn{1}{c}{\multirow{2}{*}{Method}} & \multicolumn{3}{c}{\textbf{Human Metrics}} & \multicolumn{10}{c}{\textbf{Camera Metrics}} \\
             \multicolumn{1}{c}{} & W-MPJPE$\downarrow$ & GA-MPJPE$\downarrow$ & PA-MPJPE$\downarrow$ & TE$\downarrow$ & s-TE$\downarrow$ & AE$\downarrow$ & RRA@10$\uparrow$ & RRA@15$\uparrow$ & CCA@10$\uparrow$ & CCA@15$\uparrow$ & s-CCA@10$\uparrow$ & s-CCA@15$\uparrow$ \\
            \midrule
            M0: \ours (init.) & 4.28 & 0.51 & 0.06 & 2.37 & 1.15 & 11.00 & 0.52 & 0.79 & 0.26 & 0.38 & 0.49 & 0.74 \\
            M1: $S$ \& $C$ {\tiny{w/o}} $L_{\text{Humans}}$ & 3.94 & 0.57 & 0.10 & 2.13 & 1.1 & 10.93 & 0.52 & 0.79 & 0.27 & 0.40 & 0.48 & 0.77 \\
            M2: {\tiny{w/o}} $L_{\text{Places}}$ & 1.29 & 0.24 & 0.05 & 2.82 & 0.87 & 13.02 & 0.50 & 0.73 & 0.16 & 0.24 & 0.72 & 0.88 \\
            \textbf{M3: \ours (Ours)} & \textbf{1.04} & \textbf{0.21} & \textbf{0.05} & \textbf{2.09} & \textbf{0.75} & \textbf{9.35} & \textbf{0.72} & \textbf{0.89} & \textbf{0.32} & \textbf{0.46} & \textbf{0.75} & \textbf{0.91} \\
            \midrule
            HSfM (init.) & 3.87 & 0.70 & \textbf{0.06} & 1.88 & 1.15 & 11.34 & 0.47 & 0.77 & 0.23 & 0.37 & 0.38 & 0.69 \\ 
            1 Human & 1.69 & 0.58 & \textbf{0.06} & 1.91 & 1.21 & 10.31 & 0.52 & 0.82 & 0.27 & 0.44 & 0.44 & 0.68 \\
            2 Humans & 1.66 & 0.49 & \textbf{0.06} & 1.84 & 1.10 & 9.41 & 0.62 & 0.87 & 0.33 & 0.45 & 0.53 & 0.75 \\
            3 Humans & 1.41 & \textbf{0.38} & \textbf{0.06} & 1.69 & 0.93 & 8.21 & \textbf{0.74} & \textbf{0.92} & 0.32 & 0.46 & 0.58 & 0.78 \\
            4 Humans & \textbf{1.28} & 0.39 & \textbf{0.06} & \textbf{1.52} & \textbf{0.77} & \textbf{8.11} & \textbf{0.74} & 0.90 & \textbf{0.34} & \textbf{0.53} & \textbf{0.73} & \textbf{0.90} \\
            \bottomrule
        \end{tabular}
        \caption{\textbf{Ablation study}. 
        We demonstrate the advantages of our joint optimization by removing each terms on the EgoHumans dataset. The results indicate that joint optimization is crucial for achieving a coherent reconstruction of cameras and humans. We also perform an ablation study to investigate how the number of humans included affects both camera and human pose estimation in the world coordinates by varying the number of people to include in the optimization using a subset of EgoHumans containing scenes with four people. The results reveal that the effect scales with the number of humans, highlighting the importance of leveraging multiple individuals.
        }
\label{tab:ablation}
\end{table*}

We evaluate HSfM's effectiveness in terms of human pose estimation within the world coordinate system and camera accuracy, and show qualitative results of our joint optimization on humans, scene pointmaps, and cameras.

\noindent\textbf{Evaluation Datasets: }
We evaluate on EgoHumans~\cite{khirodkar2023egohumans} and EgoExo4D~\cite{grauman2024ego}. EgoHumans is a multi-view, multi-human benchmark for human pose estimation, featuring videos of 2-4 people in real-world activities. EgoExo4D is a large-scale dataset of people performing tasks like dancing, playing music, or bike repair; see \supmat for details.

\noindent\textbf{Evaluation Metrics: }
We report metrics for humans and cameras. For people, we use the Mean Per-Joint Position Error (MPJPE). We report \textbf{W-MPJPE}, the metric measured in the world coordinate system, and \textbf{PA-MPJPE}, its Procrustes-Aligned version, measuring the local pose accuracy. We introduce Group-Aligned MPJPE, (\textbf{GA-MPJPE}), which evaluates the relative distance after Sim(3) alignment between people. All metrics are reported in meters.

For cameras, we report average camera translation error~\textbf{TE}, \ie the mean euclidean distance in meters between predicted and ground truth camera translations after SE(3) alignment. TE evaluates accuracy in the metric prediction. We also report the Sim(3) aligned version, \textbf{s-TE}. 
\textbf{AE} measures the camera Angle Error, \ie the mean Euclidean distance between predicted and ground truth camera rotation. 
\textbf{RRA} \cite{wang2023posediffusion} evaluates the Relative Rotation Accuracy by comparing the relative rotation between two predicted cameras with the corresponding ground truth.
\textbf{CCA} \cite{lin2023relpose++} assesses the Camera Center Accuracy by directly comparing the predicted and ground truth camera poses.
While Lin et al. \cite{lin2023relpose++} reported CCA only after optimal Sim(3) alignment, we provide results for two variants: the default metric computed after SE(3) alignment, and \textbf{s-CCA}, computed after Sim(3) alignment.
RRA, CCA, and s-CCA are reported for a threshold $\tau$ following the previous literature~\cite{wang2023posediffusion,wang2024dust3r,lin2023relpose++,duisterhof2024mast3r_sfm}. 
Note that W-MPJPE, TE, and CCA evaluate absolute Euclidean error, while GA-MPJPE, PA-MPJPE, s-TE, and s-CCA evaluate errors up to scale. For further details, please refer to the supplementary material.

\subsection{Results}
We compare human and camera estimation baselines on EgoHumans and EgoExo4D (\cref{tab:egohumans}), including UnCaliPose~\cite{xu2022multi}, DUSt3R~\cite{wang2024dust3r}, and MASt3R~\cite{leroy2024mast3r}. UnCaliPose jointly reconstructs humans and cameras using SfM but does not reconstruct the scene and relies on ground truth bone lengths during testing. For a fair comparison, we assume known re-identification across views, use ViTPose for UnCaliPose and \ours, and apply DUSt3R’s focal length to UnCaliPose. Since no existing approach jointly estimates people, places, and cameras at metric scale from sparse multi-view images, we also report \ours (init.), the state after our initialization in \cref{subsection:scale_init}.

On EgoHumans, \ours (init.) outperforms UnCaliPose in scale-normalized human metrics, reducing GA-MPJPE by approximately 24\% and PA-MPJPE by over 50\%. 
It nearly doubles relative rotation accuracy compared to UnCaliPose with RRA@10 and RRA@15 improvements of 86\% and 100\%, respectively. These gains show the strength of leveraging HMR2 and DUSt3R for initialization.

Our optimization further improves results: W-MPJPE drops from 4.28m in \ours (init.) to 1.04m in \ours, demonstrating the effectiveness of our approach in resolving scale ambiguity and accurately positioning humans within the world. The camera metrics also improve substantially, surpassing DUSt3R’s initial outputs and outperforming MASt3R, reducing TE from 4.97m to 2.09m and achieving about seven times better CCA@15. These results highlight the advantages of incorporating humans into the reconstruction process to achieve a consistent metric-scale world, consistent with the findings of Zhao et al.~\cite{zhao2024metric}.

Similar trends observed in EgoExo4D further validate the effectiveness of our method. \ours achieves substantially better human metrics compared to UnCaliPose, reducing W-MPJPE from 3.59m to 0.50m and PA-MPJPE from 0.13m to 0.07m. For metric-scale camera metrics, \ours outperforms \ours (init.), improving CCA@15 by 33\%. The improvements in scale-invariant camera metrics are smaller than on EgoHumans, likely due to heavy indoor occlusions affecting 2D keypoint predictions and the use of only one person in EgoExo4D. Similarly, MASt3R estimates slightly more accurate camera centers (CCA@10/15) on EgoExo4D while \ours is on-par and slightly better on the scale-invariant version (s-CCA@10/15). This is likely due to a single person being less effective for estimating scene scale; an effect we ablate in \cref{tab:ablation} on EgoHumans where we also observe a single person to be less efficient for estimating scene scale compared to more people. In contrast, EgoHumans benefits from multiple individuals, providing more 2D keypoint correspondences and strengthening optimization. Our ablation study confirms this trend.

\noindent\textbf{Ablations:} 
We validate the importance of jointly optimizing humans, scenes, and cameras in Table~\ref{tab:ablation}. The first variant, M1, detaches the gradients from the human loss to the scene and camera parameters while still optimizing all parameters. Essentially, the cameras and scene do not adjust to minimize the human losses. 
This leads to minor W-MPJPE improvement (4.28m to 3.94m) and slightly higher GA-MPJPE (0.51m to 0.57m). Camera metrics nearly stagnate (RRA@15, CCA@15) since human losses do not influence the optimization of camera and scene.
The second variant, M2, optimizes cameras and humans solely based on the human loss, excluding the scene loss. Interestingly, this significantly improves W-MPJPE (4.28m to 1.29m) and GA-MPJPE (0.57m to 0.24m), indicating accurate recovery of human world locations and relative distances. However, the camera metrics degrade considerably: CCA@15 by 36.8\%, and RRA@15 by 7.6\%. 
Without scene losses, the structure fails to anchor the cameras and overfits to human keypoints. This behavior is similar to the limitations observed in UnCaliPose, which relies solely on human keypoints for SfM. In contrast, our full method (M3) achieves the best metrics, reducing W-MPJPE and GA-MPJPE to 1.04m and 0.21m, respectively, and increasing RRA@15 by 22\% and CCA@15 by 21\%. This highlights the importance of jointly optimizing humans, scenes, and cameras to achieve a coherent and accurate metric-scale reconstruction.

We conduct an ablation study to investigate the impact of the number of humans on camera estimation and its subsequent effect on human metrics in the world coordinate system. 
As shown in the table, adding more humans consistently improves camera pose estimation, particularly by reducing camera translation error. This improvement directly contributes to achieving the lowest W-MPJPE, accurately reconstructing human locations in the world.

The results highlight that increasing the number of humans, \ie introducing more correspondences for Structure-from-Motion (SfM), strengthens the bundle adjustment process. This validates our strategy of integrating global scene optimization techniques from recent scene reconstruction methods with the traditional SfM formulation. Our approach effectively leverages 2D human keypoints as reliable correspondences and 3D human meshes as robust structures to enhance bundle adjustment and produce coherent, metric-scale reconstructions. 

Please refer to the supplementary material for an ablation study on camera/scene scale initialization and the impact of the number of cameras.

\noindent\textbf{Qualitative Results:} See~\cref{fig:teaser_mvmpw,fig:big} for qualitative results on EgoHumans and EgoExo4D and our \supmat for an \textit{in-the-wild} demo with images we captured using a minimal setup of just two cell phones. \Cref{fig:big} shows intermediate optimization steps, where in the beginning, people are floating around in mid air. After joint optimization, their feet are consistent with the environment, without any explicit contact constraints. The structure also improves as pointmaps are more coherent around the ground. Please note that the human point cloud is not used as prior for human reconstruction, \ie noisy point clouds or gaps do not directly affect the human reconstruction accuracy. Despite clear improvements in human localization, we sometimes observe scenes with slightly uneven ground planes indicating that scene reconstruction remains a challenging problem. Nonetheless, our approach improves camera metrics and scene reconstruction qualitatively (see \cref{fig:before_after_hsfm_scene}).

\section{Conclusion} 
\label{sec:conclusion}

In this work, we propose Humans and Structure from Motion, \ours, which optimizes humans, cameras, and scenes in a joint framework. We build on the success of data-driven learning in two parallel domains -- 2D/3D human reconstruction \cite{xu2022vitpose,goel2023humans} and scene reconstruction \cite{wang2024dust3r}. However, neither of these approaches is able to reconstruct humans, scenes, and cameras coherently. Our experiments verify the synergy between these three elements --- integrating human reconstruction into the classic SfM task not only properly places people in the world, but also significantly improves camera  pose accuracy. Despite promising results, as optimization-based framework our approach can be sensitive to hyper-parameters. In future work, it would be interesting to explore this synergy in a feed-forward framework with integrated re-ID or leverage recent work like MONSt3R~\cite{zhang2024monst3r} to extend our insights to videos.

\section*{Acknowledgements} This project is supported in part by DARPA No. HR001123C0021, IARPA DOI/IBC No. 140D0423C0035, NSF:CNS-2235013, ONR MURI N00014-21-1-2801, Bakar Fellows, and Bair Sponsors. The views and conclusions contained herein are those of the authors and do not represent the official policies or endorsements of these institutions. We also thank Chung Min Kim for her critical reviews on this paper and Junyi Zhang for his valuable insights on the method.

%% file: sections/04_Appendix_CameraReady.tex
\begin{table*}
    \footnotesize
    \centering
    \setlength{\tabcolsep}{0.8pt}
        \begin{tabular}{lccccccccccccc}
            \toprule
             \multicolumn{1}{c}{\multirow{2}{*}{}} & \multicolumn{3}{c}{\textbf{Human Metrics}} & \multicolumn{10}{c}{\textbf{Camera Metrics}} \\
             \multicolumn{1}{c}{} & W-MPJPE$\downarrow$ & GA-MPJPE$\downarrow$ & PA-MPJPE$\downarrow$ & TE$\downarrow$ & s-TE$\downarrow$ & AE$\downarrow$ & RRA@10$\uparrow$ & RRA@15$\uparrow$ & CCA@10$\uparrow$ & CCA@15$\uparrow$ & s-CCA@10$\uparrow$ & s-CCA@15$\uparrow$ \\
            \midrule
                        S1: $\alpha = 1.0$ & 11.89 & 0.85 & 0.09 & 6.42 & 5.50 & 121.88 & 0.01 & 0.01 & 0.01 & 0.02 & 0.01 & 0.05 \\
            S2: $\alpha = 100.0$ & 1.94 & 0.22 & 0.06 & 2.17 & 1.11 & 15.00 & 0.68 & 0.82 & 0.31 & 0.45 & 0.68 & 0.83 \\
                        \textbf{S3: \ours (Ours)} & \textbf{1.04} & \textbf{0.21} & \textbf{0.05} & \textbf{2.09} & \textbf{0.75} & \textbf{9.35} & \textbf{0.72} & \textbf{0.89} & \textbf{0.32} & \textbf{0.46} & \textbf{0.75} & \textbf{0.91} \\

            \midrule  

            2 Cam. HSfM (init) & 3.73 & 0.42 & 0.06 & 1.53 & - & 9.81 & 0.41 & 0.87 & 0.08 & 0.12 & - & - \\ 
            \rowcolor[gray]{0.9} 2 Cam. HSfM (Ours) & 2.63 & 0.26 & \textbf{0.05} & \textbf{0.39} & - & 10.37 & 0.41 & \textbf{0.91} & \textbf{0.48} & \textbf{0.68} & - & - \\ 
            4 Cam. HSfM (init) & 4.26 & 0.51 & 0.06 & 2.36 & 1.14 & 10.96 & 0.52 & 0.79 & 0.26 & 0.38 & 0.49 & 0.74 \\
            \rowcolor[gray]{0.9} 4 Cam. HSfM (Ours) & 1.15 & 0.27 & 0.06 & 2.00 & \textbf{0.71} & 8.92 & 0.68 & 0.88 & 0.35 & 0.50 & \textbf{0.78} & \textbf{0.93} \\
            8 Cam. HSfM (init) & 5.06 & 0.53 & 0.06 & 2.36 & 0.96 & 7.61 & 0.71 & 0.87 & 0.25 & 0.40 & 0.65 & 0.88 \\
            \rowcolor[gray]{0.9} 8 Cam. HSfM (Ours) & \textbf{1.00} & \textbf{0.19} & \textbf{0.05} & 1.97 & 0.90 & \textbf{7.41} & \textbf{0.76} & 0.90 & 0.41 & 0.54 & 0.72 & 0.88 \\
            \bottomrule
        \end{tabular}
        \caption{\textbf{Ablation on the number of input view cameras.} We evaluate the performance of HSfM by varying the number of input view cameras and assessing human reconstruction and camera pose estimation in the world coordinate frame. The experiments are conducted on EgoHumans, excluding samples without ground truth camera poses for all views in the specified combinations (2, 4, and 8). Compared to the initialization, our joint optimization improves all human pose and camera pose metrics, regardless of the number of input cameras. We do not report the scaled version of camera translation errors for the 2-camera cases, as the predictions become identical to the ground truth camera translations after scale alignment.
        }
\label{tab:ablation_num_views_suppl}
\end{table*}

This is the supplementary material for our main paper ``Reconstructing People, Places, and Cameras". We provide additional qualitative results (\cref{sec:additional_qual}), including in-the-wild examples, a discussion of our approach's limitations (\cref{sec:discussion}), ablation studies on the scale initialization and input views (\cref{sec:ablations}), evaluation details (\cref{sec:eval_detail}), implementation detail (\cref{section:implementation}), and additional related work (\cref{section:additional_related_work}). 


\begin{figure*}[t]
    \centering
    \includegraphics[width=\linewidth,clip,trim=0cm 0cm 0cm 0cm]{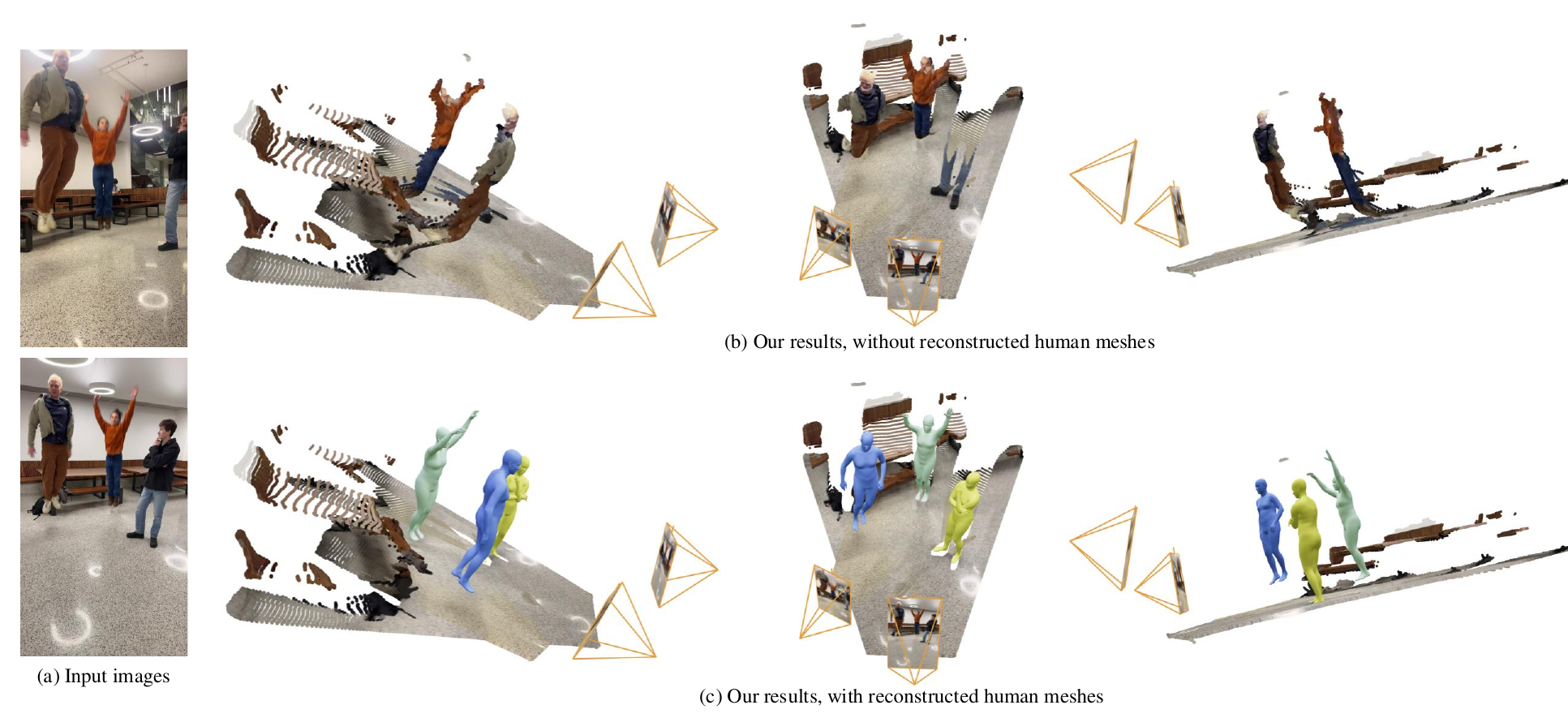}
    \caption[]{\textbf{Qualitative results \textit{in the wild}.} 
    We show reconstructions on \textit{in-the-wild} images taken with two smartphones (a), demonstrating the reconstruction of humans and scenes. Unlike previous works~\cite{zou2020reducing,ye2023decoupling}, which adopt human-scene contact priors that hinder generalization to scenarios without ground foot contact, \ours recovers accurate world locations of the human meshes that are coherent with the static scene structure. The use of humans in our framework (c) not only serves as a reliable initialization for 3D structure in the SfM formulation but also provides more faithful and complete information about people in the world, which a noisy human point cloud (b) cannot offer. For visualization purposes, the human point cloud is removed using SAM2~\cite{ravi2024sam2}.
    }
    \label{fig:inthewild}
\end{figure*}

\begin{figure*}
    \centering
    \begin{subfigure}{0.989\textwidth}
        \centering
        \includegraphics[width=0.965\textwidth]{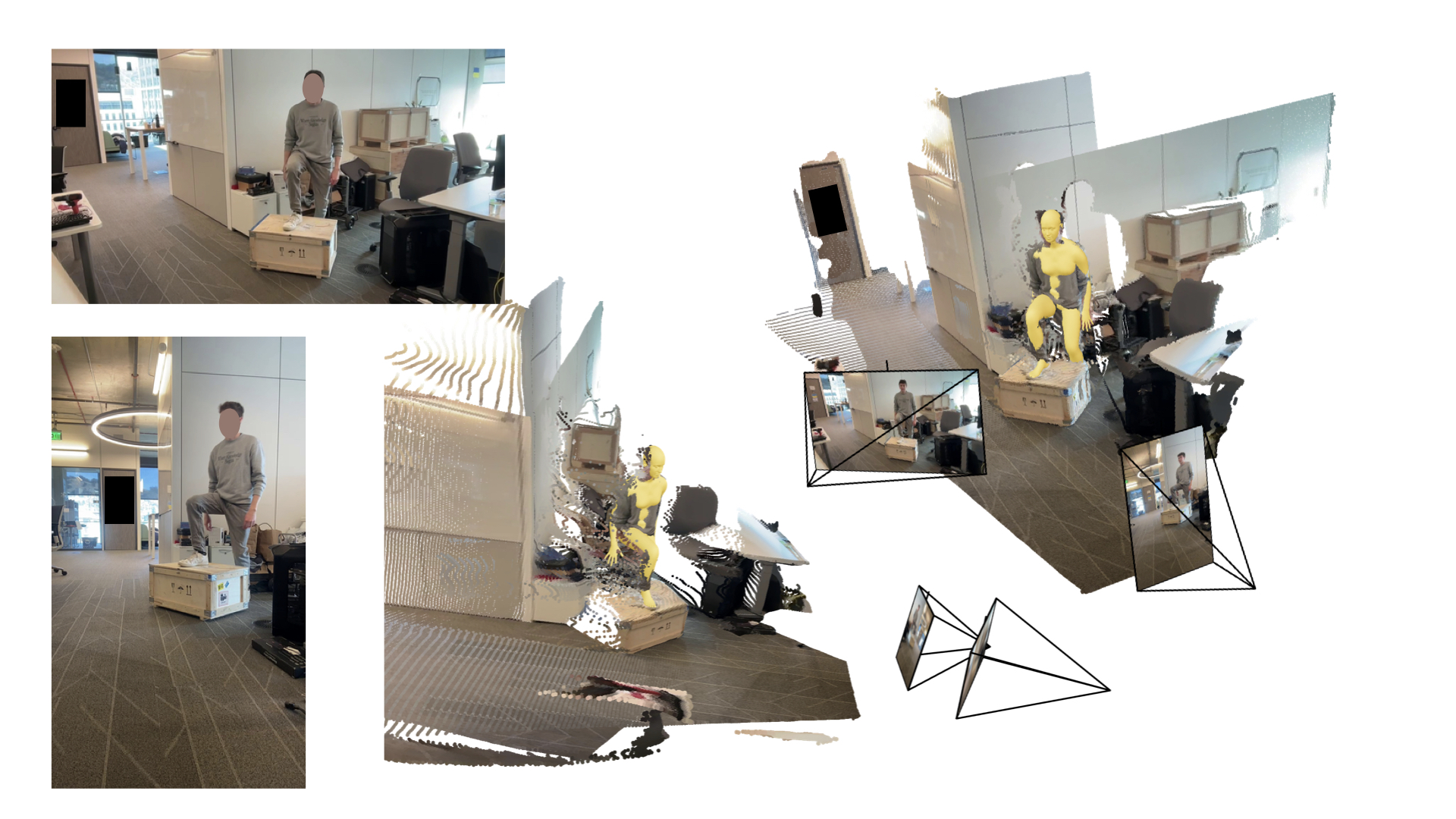}
    \end{subfigure}
    \vspace{1cm}
    \begin{subfigure}{0.989\textwidth}
        \centering
        \includegraphics[width=0.965\textwidth]{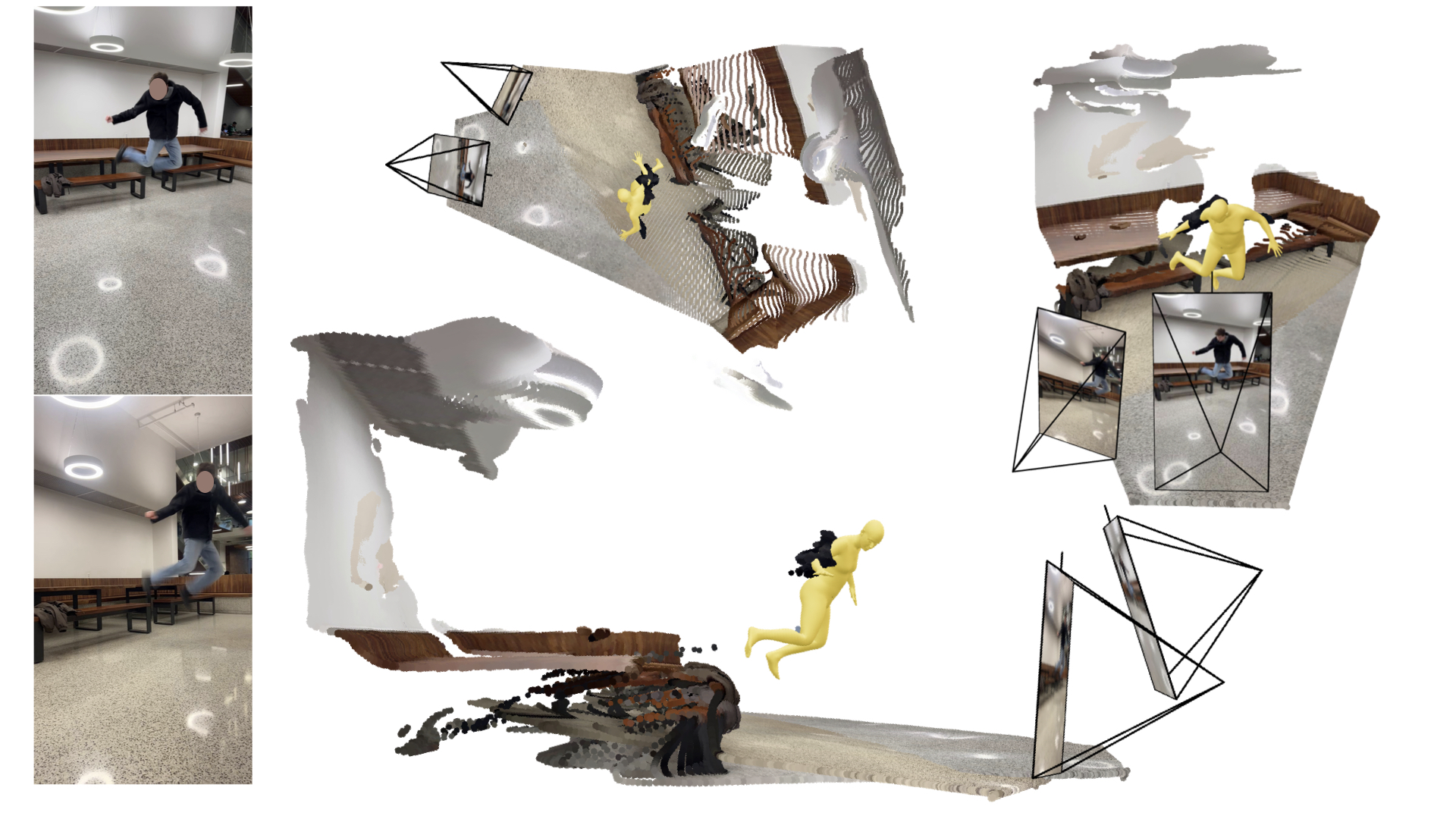}
    \end{subfigure}
    \caption{\textbf{Qualitative results \textit{in the wild}.} We show reconstructions on \textit{in-the-wild} images taken with two cell phones and the reconstruction of humans and scene. Our method places people in the world and reconstructs accurate human-scene contact, \eg between the person's right foot and box.}
    \label{fig:stacked_images_demo_02}
    \vspace{2cm}
\end{figure*}

\section{Additional Qualitative Results}
\label{sec:additional_qual}
\textbf{In-the-wild demo.} We first present HSfM's reconstruction results on images captured by two cell phones in Figures~\ref{fig:inthewild} and ~\ref{fig:stacked_images_demo_02}. 
The data was captured using a minimal setup consisting of two cell phones, two tripods, and the Riverside app\footnote{https://riverside.fm/} for straightforward time synchronization. Despite this simple setup, our multi-view optimization algorithm successfully handles challenging scenes, such as individuals jumping, without relying on any heuristic contact priors and small data-driven motion priors that previous works~\cite{ye2023decoupling,shin2024wham,zhao2024synchmr} use.

\begin{figure*}
    \centering
    \includegraphics[width=0.9\textwidth]{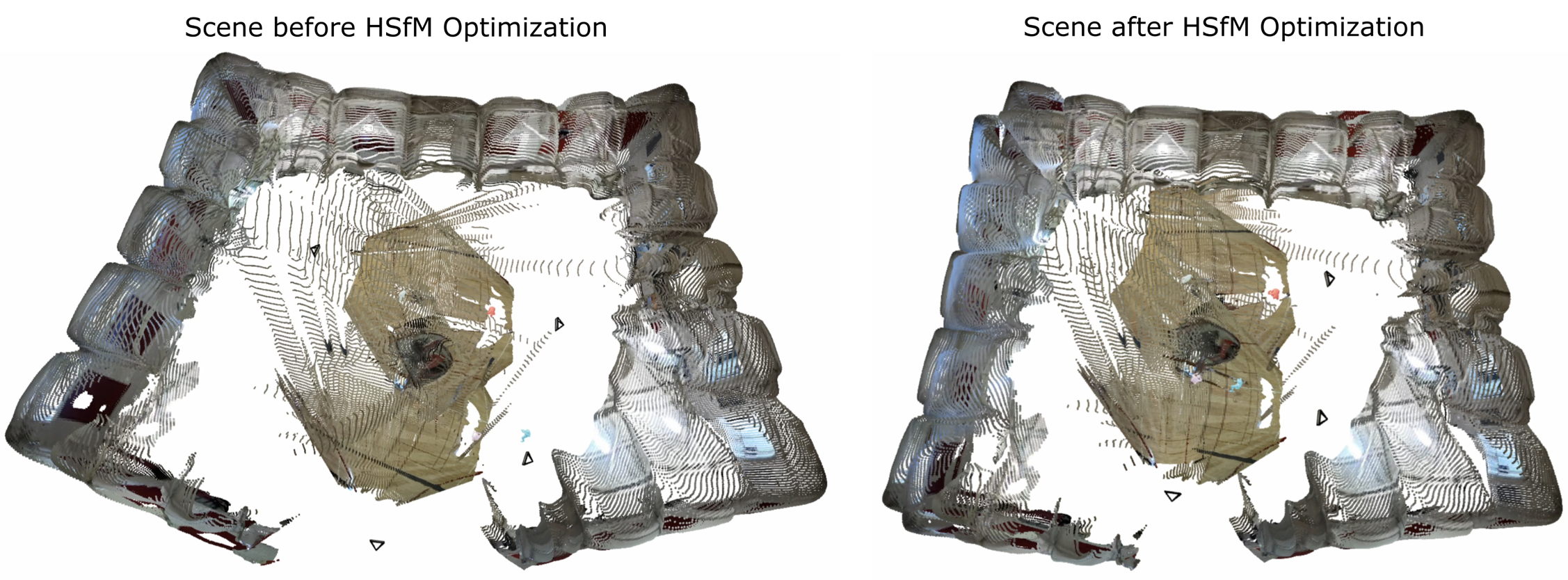}
    \caption{\textbf{Qualitative result of HSfM reconstruction.} Top view of scene from EgoHumans before and after HSfM optimization.}
    \label{fig:before_after_hsfm_scene}
\end{figure*}

\textbf{Benchmark evaluation.} We provide additional qualitative results on EgoExo4D \cite{grauman2024ego} in \Cref{fig:stacked_images_01}, showing challenging scenes such as kitchens, humans interacting with objects (e.g., playing the piano), and sports activities like soccer. In \Cref{fig:stacked_images_04}, we display further results on EgoHumans \cite{khirodkar2023egohumans}, demonstrating HSfM reconstructions of multiple people interacting, such as fencing. \Cref{fig:before_after_hsfm_scene} show a scene from EgoHumans before and after HSfM optimization.

\section{Discussion}
\label{sec:discussion}

Our goal is to study the mutual benefits of jointly reconstructing humans, scenes, and cameras.
To this end, we assume that the re-identification of people across camera views is known, because misidentified individuals can introduce spurious effects, disrupting the optimization process. 
Please note that this limitation also applies to UnCaliPose \cite{xu2019denserac}. 
To ensure a fair comparison and maintain focus on the core objectives of this study, we rely on ground-truth identities in both our approach and UnCaliPose in the main text.

Since re-identification may not be available at test time and manual identification is cumbersome, we tested the feasibility of automating the re-identification process using the re-identification module of UnCaliPose on the EgoHumans dataset \cite{khirodkar2023egohumans}. For re-identification, UnCaliPose solves a constrained clustering optimization problem, assuming a known number of people in the scene and utilizing re-identification features extracted by an off-the-shelf re-identification network~\cite{luo2019bag}. The re-identification process achieves an accuracy of 51.22\% on EgoHumans. The main failure mode occurs with individuals wearing uniforms (e.g., tennis sequences: 12.04\% accuracy, volleyball sequences: 25.71\% accuracy), where appearance features are difficult to distinguish.

These findings indicate that manual re-identification remains necessary for accurate multi-view reconstruction of the world, including humans. Fortunately, modern tools like LabelMe\footnote{https://github.com/wkentaro/labelme} simplify this process. Looking ahead, we anticipate that ongoing advancements in large-scale data-model paradigms will significantly improve performance in multi-view re-identification. These advancements may include robust appearance feature matching~\cite{oquab2023dinov2} and the use of geometric similarities, such as human pose and location~\cite{rajasegaran2022tracking,goel2023humans}.

While our method achieves good quantitative results, we observe a few failure cases stemming from preprocessing errors in reconstruction and detections. The most common issues arise from erroneous initial camera estimates generated by the scene reconstruction-based SfM\cite{wang2024dust3r}, particularly in scenes with limited structure, insufficient overlap between images, or large areas affected by radial distortion. In instances where DUSt3R\cite{wang2024dust3r} fails to detect any cameras, we rely on human-centric camera poses to initialize the optimization. Another source of error involves missing or highly inaccurate keypoint detections, which can occur under conditions of heavy occlusion or poor lighting. In such cases, our method estimates frame-specific cameras solely based on pixel data, without incorporating human constraints. 
Despite these occasional errors, we find DUSt3R, ViTPose \cite{xu2022vitpose}, and HMR2.0 \cite{goel2023humans} to exhibit remarkable robustness across a wide range of challenging scenarios.

\section{Additional Ablation Studies}
\label{sec:ablations}

We analyze the effect of different scale initializations to validate the superiority of the human-centric scaling introduced in Section~\ref{subsection:scale_init} in Table~\ref{tab:ablation_num_views_suppl}. Without scale initialization ($\alpha = 1.0$), where we directly use the raw DUSt3R~\cite{wang2024dust3r} scene and camera pose outputs as input to our optimization, the W-MPJPE is 11.89m, whereas ours is 1.04m. Additionally, the high metric-scale camera translation errors, such as 6.42m TE, and extremely low RRA values, demonstrate the necessity of proper initialization. This error occurs because the raw camera and scene outputs have a significantly smaller scale than the real world due to their scale normalization during training.

Choosing a large scale value ($\alpha = 100$) generally covers the real-world capture scene sufficiently but does not perform as well as our human-centric scaling approach (W-MPJPE: 1.94m vs. 1.04m). The camera metrics are also worse than ours (e.g., RRA values are 5–7\% lower than ours). This implies that, without proper scaling, the optimization is prone to failure due to poor initialization and local minima problems.

One common local minimum observed was humans being placed behind the camera while still reprojecting to the correct pixel locations. To address this, we increased the scale $\alpha$ until all humans were placed in front of all cameras, ensuring positive depth values in all camera coordinate systems. While this initialization produced similar quantitative results in successful cases, it completely failed for 2\% of samples, demonstrating that naive initialization approaches are not reliable.

Next, we vary the number of cameras to evaluate the robustness of our method. We tested 2, 4, and 8 input view cameras. As indicated in the table, our joint optimization consistently improves all metrics, regardless of the number of input view cameras. With only 2 cameras, W-MPJPE is 2.63m and GA-MPJPE is 0.26m, indicating accurate human placement in the world. The consistently better camera results compared to the initialization further validate the benefit of incorporating humans into the traditional SfM formulation. The robustness of our method is further demonstrated qualitatively in Figure~\ref{fig:stacked_images_demo_02}, where the data is captured by two cameras in in-the-wild scenes.

\section{Evaluation Details}
\label{sec:eval_detail}

\subsection{Evaluation Metrics}
In this section, we provide additional details about our human pose and camera metrics.

\noindent \textbf{W-MPJPE} describes the mean per-joint position error, measured in the world frame. To bring predicted human meshes into the ground-truth's world coordinate system, we use an SE(3) rigid alignment from the estimated camera positions to the ground-truth camera positions.

\noindent \textbf{PA-MPJPE} describes the Procrustes-aligned variant of MPJPE, which measures position errors after Sim(3) alignment of joints for each human.
This metric evaluates local pose accuracy in a way that is not dependent on camera position estimates or human body scale.

\noindent \textbf{GA-MPJPE} evaluates group-aligned joint position errors, computed after Sim(3) alignment for all humans in a scene. This measures people relative to each other, without considering the scene or camera positions.

\noindent \textbf{TE} measures the mean Euclidean distance between predicted and ground truth camera positions, after SE(3) alignment. TE evaluates metric accuracy of camera positions. 

\noindent \textbf{s-TE} is the scale-aligned version of TE, where we preprocess positions with Sim(3) instead of SE(3) alignment. This measures scale-invariant errors for estimated cameras.

\noindent \textbf{AE} measures the average Angle Error between camera pairs. We compute relative orientations for each pair of cameras in a scene. We then measure the difference between ground-truth and predicted pairwise orientations, convert to degrees, and average.

\noindent\textbf{CCA}~\cite{lin2023relpose++} measures the Camera Center Accuracy, after the SE(3) alignment process used for TE.
CCA@$\tau$ is the proportion of camera positions with absolute error within $\tau$\% of the overall scene scale.
Following existing work, we compute the scene scale as the furthest distance between a ground-truth camera and the centroid.

\noindent \textbf{s-CCA} measures the the scale-aligned version of CCA, after the Sim(3) alignment process used for s-TE.

\noindent \textbf{RRA}~\cite{wang2023posediffusion} measures the Relative Rotation Accuracy of camera estimates, computed using the same camera pairs as AE.
RRA@$\tau$ is the proportion of pairwise camera orientations with angular error of $\tau$ degrees or lower.

\subsection{Evaluation Datasets}

\noindent\textbf{EgoHumans:}
In the main text's tables and Table~\ref{tab:ablation_num_views_suppl}'s 4 view case, we used the following camera configurations for each sequence:  

\begin{itemize}
    \item For \texttt{01\_tagging} sequences: \texttt{camera 1}, \texttt{camera 4}, \texttt{camera 6}, and \texttt{camera 8}.
    \item For \texttt{02\_lego} sequences: \texttt{camera 2}, \texttt{camera 3}, \texttt{camera 4}, and \texttt{camera 6}.
    \item For \texttt{03\_fencing} sequences: \texttt{camera 4}, \texttt{camera 5}, \texttt{camera 10}, and \texttt{camera 13}.
    \item For \texttt{04\_basketball} sequences: \texttt{camera 1}, \texttt{camera 3}, \texttt{camera 4}, and \texttt{camera 8}.
    \item For \texttt{05\_volleyball} sequences: \texttt{camera 2}, \texttt{camera 4}, \texttt{camera 8}, and \texttt{camera 11}.
    \item For \texttt{06\_badminton} sequences: \texttt{camera 1}, \texttt{camera 2}, \texttt{camera 5}, and \texttt{camera 7}.
    \item For \texttt{07\_tennis} sequences: \texttt{camera 4}, \texttt{camera 9}, \texttt{camera 12}, and \texttt{camera 20}.
\end{itemize}

In Table~\ref{tab:ablation_num_views_suppl}'s 2 view case, we used the following camera configurations for each sequence:  

\begin{itemize}
    \item For \texttt{01\_tagging} sequences: \texttt{camera 1} and \texttt{camera 2}.
    \item For \texttt{02\_lego} sequences: \texttt{camera 3} and \texttt{camera 5}.
    \item For \texttt{03\_fencing} sequences: \texttt{camera 5} and \texttt{camera 13}.
    \item For \texttt{04\_basketball} sequences: \texttt{camera 2} and \texttt{camera 7}.
    \item For \texttt{05\_volleyball} sequences: \texttt{camera 6} and \texttt{camera 12}.
    \item For \texttt{06\_badminton} sequences: \texttt{camera 5} and \texttt{camera 7}.
    \item For \texttt{07\_tennis} sequences: \texttt{camera 9} and \texttt{camera 12}.
\end{itemize}

In Table~\ref{tab:ablation_num_views_suppl}'s 8 view case, we used the following camera configurations for each sequence:  

\begin{itemize}
    \item For \texttt{01\_tagging} sequences: all 8 available cameras.
    \item For \texttt{02\_lego} sequences: all 8 available cameras.
    \item For \texttt{03\_fencing} sequences: \texttt{camera 1}, \texttt{camera 3}, \texttt{camera 5}, \texttt{camera 7}, \texttt{camera 9}, \texttt{camera 11}, \texttt{camera 13}, \texttt{camera 15}.
    \item For \texttt{04\_basketball} sequences: all 8 available cameras.
    \item For \texttt{05\_volleyball} sequences: \texttt{camera 1}, \texttt{camera 3}, \texttt{camera 5}, \texttt{camera 7}, \texttt{camera 9}, \texttt{camera 11}, \texttt{camera 13}, \texttt{camera 15}.
    \item For \texttt{06\_badminton} sequences: \texttt{camera 1}, \texttt{camera 3}, \texttt{camera 5}, \texttt{camera 7}, \texttt{camera 9}, \texttt{camera 11}, \texttt{camera 13}, \texttt{camera 15}.
    \item For \texttt{07\_tennis} sequences: \texttt{camera 1}, \texttt{camera 3}, \texttt{camera 5}, \texttt{camera 7}, \texttt{camera 9}, \texttt{camera 11}, \texttt{camera 13}, \texttt{camera 15}.
\end{itemize}

\noindent\textbf{EgoExo4D:} 
EgoExo4D scenes are typically captured using four to six RGB cameras and an egocentric device (Aria glasses). For our experiments and the baselines, we use only the RGB images from sequences with correct re-identification. Sequences containing ego-centric RGB views, such as helmet-mounted cameras, are excluded. We evaluate 182 videos from the validation set, sampling one random frame per video. The videos include ground-truth annotations for human poses, locations, and camera parameters. We evaluate on the following takes/frames:
\texttt{ cmu\_soccer06\_3/1426}\\
\texttt{ cmu\_soccer12\_2/6807}\\
\texttt{ cmu\_soccer16\_2/6373}\\
\texttt{ georgiatech\_bike\_06\_12/170}\\
\texttt{ georgiatech\_bike\_06\_2/97}\\
\texttt{ georgiatech\_bike\_06\_6/74}\\
\texttt{ georgiatech\_bike\_06\_8/15}\\
\texttt{ georgiatech\_bike\_07\_10/28}\\
\texttt{ georgiatech\_bike\_07\_12/38}\\
\texttt{ georgiatech\_bike\_07\_2/97}\\
\texttt{ georgiatech\_bike\_07\_4/46}\\
\texttt{ georgiatech\_bike\_07\_6/67}\\
\texttt{ georgiatech\_bike\_07\_8/138}\\
\texttt{ georgiatech\_bike\_14\_12/593}\\
\texttt{ georgiatech\_bike\_14\_2/1214}\\
\texttt{ georgiatech\_bike\_14\_6/575}\\
\texttt{ georgiatech\_bike\_14\_8/97}\\
\texttt{ georgiatech\_bike\_15\_2/1508}\\
\texttt{ georgiatech\_bike\_15\_4/844}\\
\texttt{ georgiatech\_bike\_15\_6/1103}\\
\texttt{ georgiatech\_bike\_15\_8/3153}\\
\texttt{ georgiatech\_bike\_16\_2/882}\\
\texttt{ georgiatech\_bike\_16\_6/3031}\\
\texttt{ georgiatech\_bike\_16\_8/1274}\\
\texttt{ georgiatech\_covid\_02\_10/2227}\\
\texttt{ georgiatech\_covid\_02\_12/6974}\\
\texttt{ georgiatech\_covid\_02\_14/2926}\\
\texttt{ georgiatech\_covid\_02\_2/67}\\
\texttt{ georgiatech\_covid\_02\_4/67}\\
\texttt{ georgiatech\_covid\_04\_10/999}\\
\texttt{ georgiatech\_covid\_04\_12/6160}\\
\texttt{ georgiatech\_covid\_04\_4/2996}\\
\texttt{ georgiatech\_covid\_04\_6/4528}\\
\texttt{ georgiatech\_covid\_06\_2/47}\\
\texttt{ georgiatech\_covid\_06\_4/64}\\
\texttt{ georgiatech\_covid\_18\_10/5524}\\
\texttt{ georgiatech\_covid\_18\_12/3457}\\
\texttt{ georgiatech\_covid\_18\_2/2413}\\
\texttt{ georgiatech\_covid\_18\_4/3534}\\
\texttt{ georgiatech\_covid\_18\_6/4389}\\
\texttt{ georgiatech\_covid\_18\_8/458}\\
\texttt{ iiith\_cooking\_59\_2/7795}\\
\texttt{ iiith\_cooking\_64\_2/298}\\
\texttt{ iiith\_cooking\_89\_6/1177}\\
\texttt{ iiith\_cooking\_90\_4/1383}\\
\texttt{ iiith\_soccer\_015\_2/1610}\\
\texttt{ nus\_cpr\_12\_1/1338}\\
\texttt{ nus\_cpr\_12\_2/76}\\
\texttt{ sfu\_basketball012\_10/774}\\
\texttt{ sfu\_basketball012\_12/399}\\
\texttt{ sfu\_basketball012\_2/945}\\
\texttt{ sfu\_basketball012\_3/1506}\\
\texttt{ sfu\_basketball012\_4/66}\\
\texttt{ sfu\_basketball012\_6/526}\\
\texttt{ sfu\_basketball012\_7/1581}\\
\texttt{ sfu\_basketball012\_8/329}\\
\texttt{ sfu\_basketball016\_2/247}\\
\texttt{ sfu\_basketball\_04\_8/209}\\
\texttt{ sfu\_basketball\_05\_22/1902}\\
\texttt{ sfu\_basketball\_05\_26/29}\\
\texttt{ sfu\_basketball\_09\_11/32}\\
\texttt{ sfu\_basketball\_09\_12/1114}\\
\texttt{ sfu\_cooking028\_12/1049}\\
\texttt{ sfu\_cooking\_007\_7/77}\\
\texttt{ sfu\_cooking\_008\_3/4164}\\
\texttt{ sfu\_cooking\_008\_5/3559}\\
\texttt{ sfu\_covid\_004\_2/2828}\\
\texttt{ sfu\_covid\_004\_4/5360}\\
\texttt{ sfu\_covid\_008\_16/1595}\\
\texttt{ unc\_basketball\_02-24-23\_01\_3/84}\\
\texttt{ unc\_basketball\_02-24-23\_02\_10/466}\\
\texttt{ unc\_basketball\_02-24-23\_02\_11/927}\\
\texttt{ unc\_basketball\_03-30-23\_02\_10/45}\\
\texttt{ unc\_basketball\_03-30-23\_02\_14/7}\\
\texttt{ unc\_basketball\_03-30-23\_02\_15/40}\\
\texttt{ unc\_basketball\_03-30-23\_02\_17/9}\\
\texttt{ unc\_basketball\_03-30-23\_02\_18/20}\\
\texttt{ unc\_basketball\_03-30-23\_02\_19/7}\\
\texttt{ unc\_basketball\_03-30-23\_02\_4/107}\\
\texttt{ unc\_basketball\_03-30-23\_02\_5/25}\\
\texttt{ unc\_basketball\_03-30-23\_02\_7/1141}\\
\texttt{ uniandes\_basketball\_001\_23/768}\\
\texttt{ uniandes\_basketball\_001\_24/1386}\\
\texttt{ uniandes\_basketball\_001\_26/146}\\
\texttt{ uniandes\_basketball\_001\_27/439}\\
\texttt{ uniandes\_basketball\_003\_38/32}\\
\texttt{ uniandes\_basketball\_004\_23/369}\\
\texttt{ uniandes\_basketball\_004\_44/261}\\
\texttt{ uniandes\_basketball\_004\_45/667}\\
\texttt{ uniandes\_dance\_002\_11/201}\\
\texttt{ uniandes\_dance\_002\_2/439}\\
\texttt{ uniandes\_dance\_008\_29/276}\\
\texttt{ uniandes\_dance\_008\_30/166}\\
\texttt{ uniandes\_dance\_008\_31/31}\\
\texttt{ uniandes\_dance\_008\_32/11}\\
\texttt{ uniandes\_dance\_008\_33/1105}\\
\texttt{ uniandes\_dance\_008\_34/753}\\
\texttt{ uniandes\_dance\_008\_35/607}\\
\texttt{ uniandes\_dance\_008\_36/1045}\\
\texttt{ uniandes\_dance\_008\_37/913}\\
\texttt{ uniandes\_dance\_008\_38/706}\\
\texttt{ uniandes\_dance\_016\_10/841}\\
\texttt{ uniandes\_dance\_016\_11/279}\\
\texttt{ uniandes\_dance\_016\_12/932}\\
\texttt{ uniandes\_dance\_016\_13/453}\\
\texttt{ uniandes\_dance\_016\_14/951}\\
\texttt{ uniandes\_dance\_016\_30/577}\\
\texttt{ uniandes\_dance\_016\_31/1709}\\
\texttt{ uniandes\_dance\_016\_32/377}\\
\texttt{ uniandes\_dance\_016\_33/1158}\\
\texttt{ uniandes\_dance\_016\_36/1247}\\
\texttt{ uniandes\_dance\_016\_37/145}\\
\texttt{ uniandes\_dance\_016\_38/1416}\\
\texttt{ uniandes\_dance\_016\_39/399}\\
\texttt{ uniandes\_dance\_016\_3/1239}\\
\texttt{ uniandes\_dance\_016\_42/1406}\\
\texttt{ uniandes\_dance\_016\_43/1271}\\
\texttt{ uniandes\_dance\_016\_44/1268}\\
\texttt{ uniandes\_dance\_016\_45/838}\\
\texttt{ uniandes\_dance\_016\_6/1361}\\
\texttt{ uniandes\_dance\_016\_7/1040}\\
\texttt{ uniandes\_dance\_016\_8/1488}\\
\texttt{ uniandes\_dance\_017\_6/1592}\\
\texttt{ uniandes\_dance\_019\_17/1003}\\
\texttt{ uniandes\_dance\_019\_18/509}\\
\texttt{ uniandes\_dance\_019\_19/1537}\\
\texttt{ uniandes\_dance\_019\_20/1089}\\
\texttt{ uniandes\_dance\_019\_22/81}\\
\texttt{ uniandes\_dance\_019\_24/484}\\
\texttt{ uniandes\_dance\_019\_25/183}\\
\texttt{ uniandes\_dance\_019\_26/1814}\\
\texttt{ uniandes\_dance\_019\_27/283}\\
\texttt{ uniandes\_dance\_019\_28/1411}\\
\texttt{ uniandes\_dance\_019\_46/412}\\
\texttt{ uniandes\_dance\_019\_47/790}\\
\texttt{ uniandes\_dance\_019\_49/1617}\\
\texttt{ uniandes\_dance\_019\_51/481}\\
\texttt{ uniandes\_dance\_019\_52/875}\\
\texttt{ uniandes\_dance\_019\_54/766}\\
\texttt{ uniandes\_dance\_019\_55/679}\\
\texttt{ uniandes\_dance\_019\_56/561}\\
\texttt{ uniandes\_dance\_019\_57/1073}\\
\texttt{ uniandes\_dance\_019\_58/192}\\
\texttt{ uniandes\_dance\_024\_11/1619}\\
\texttt{ uniandes\_dance\_024\_12/104}\\
\texttt{ uniandes\_dance\_024\_13/1419}\\
\texttt{ uniandes\_dance\_024\_14/1180}\\
\texttt{ uniandes\_dance\_024\_15/378}\\
\texttt{ uniandes\_dance\_024\_16/1569}\\
\texttt{ uniandes\_dance\_024\_17/1317}\\
\texttt{ uniandes\_dance\_024\_45/844}\\
\texttt{ uniandes\_dance\_024\_47/732}\\
\texttt{ uniandes\_dance\_024\_48/261}\\
\texttt{ uniandes\_dance\_024\_49/325}\\
\texttt{ upenn\_0706\_Dance\_4\_2/2512}\\
\texttt{ upenn\_0706\_Dance\_4\_3/1277}\\
\texttt{ upenn\_0706\_Dance\_4\_4/1670}\\
\texttt{ upenn\_0706\_Dance\_4\_5/1904}\\
\texttt{ upenn\_0713\_Dance\_3\_2/164}\\
\texttt{ upenn\_0713\_Dance\_3\_3/586}\\
\texttt{ upenn\_0713\_Dance\_3\_4/0}\\
\texttt{ upenn\_0713\_Dance\_3\_5/243}\\
\texttt{ upenn\_0713\_Dance\_4\_2/125}\\
\texttt{ upenn\_0713\_Dance\_4\_3/1280}\\
\texttt{ upenn\_0713\_Dance\_4\_4/308}\\
\texttt{ upenn\_0713\_Dance\_4\_5/262}\\
\texttt{ upenn\_0713\_Dance\_5\_4/238}\\
\texttt{ upenn\_0713\_Dance\_5\_6/2534}\\
\texttt{ upenn\_0721\_Piano\_1\_2/140}\\
\texttt{ upenn\_0721\_Piano\_1\_3/648}\\
\texttt{ upenn\_0722\_Piano\_1\_2/83}\\
\texttt{ upenn\_0727\_Partner\_Dance\_3\_1\_2/62}\\
\texttt{ utokyo\_pcr\_2001\_29\_2/5799}\\
\texttt{ utokyo\_pcr\_2001\_29\_4/3491}\\
\texttt{ utokyo\_pcr\_2001\_29\_6/550}\\
\texttt{ utokyo\_pcr\_2001\_30\_2/2121}\\
\texttt{ utokyo\_pcr\_2001\_30\_4/1696}\\
\texttt{ utokyo\_pcr\_2001\_32\_2/6641}\\
\texttt{ utokyo\_pcr\_2001\_32\_4/6048}\\
\texttt{ utokyo\_soccer\_8000\_43\_2/3262}\\
\texttt{ utokyo\_soccer\_8000\_43\_4/3472}\\
\texttt{ utokyo\_soccer\_8000\_43\_6/2781}\\

\input{sections/MVMPW/figtex/supp_qual}

\section{Implementation Details}
\label{section:implementation} 
Given sparse-view images, \ours jointly estimates SMPL-X~\cite{Pavlakos2019_smplifyx} parameters for humans, scene pointmaps, and camera poses (rotation and translation), in the world coordinate frame. The SMPL-X parameters for humans are initialized using predictions from HMR2~\cite{goel2023humans} converted to \smplx following the conversion procedure in \cite{mueller2023buddi}. Scene pointmaps and camera parameters are initialized with estimates from DUSt3R~\cite{wang2024dust3r}. 
We use Adam~\cite{adam} optimizer and set the number of optimization steps proportional to the scene scale with a minimum of 500 steps. This allows sufficient time to accurately determine scene scale and camera poses and people's location. The learning rate is set to 0.015 with a linear reduction schedule. To tune hyperparameters, we use the first frame (4 cameras) of sequences \texttt{01\_tagging} and \texttt{04\_basketball} of EgoHumans as these scene encompass a good range of scene scales.

\section{Additional related work}
\label{section:additional_related_work}
\textbf{Monocular Human Mesh Reconstruction.} 
Most methods estimate 3D humans in the camera coordinate system for a single person \cite{Kanazawa2018_hmr,Pavlakos2019_smplifyx,goel2023humans,dwivedi2024tokenhmr} or for multiple people with depth estimation \cite{sun2022putting,baradel2024multi,zhang2022egobody}. Recent works jointly estimate human and camera motion in the world coordinate frame \cite{yuan2022glamr,yu2023trace,ye2023decoupling,shin2024wham,shen2024world,wang2024tram,liu20214d,zhao2024synchmr}. They leverage temporal dynamics from video sequences to improve reconstruction quality over time. While single-view reconstruction methods are valuable for their minimal input requirements, they often suffer from ambiguity, especially due to occlusion. Our approach leverages multi-view data to enhance reconstruction accuracy and integrates scene context, providing a more detailed and reliable reconstruction of multi-person interactions within their environment.

%% file: sections/MVMPW/figtex/supp_qual.tex
\begin{figure*}[htbp]
    \centering
    \begin{subfigure}[t]{0.95\textwidth}
        \centering
        \includegraphics[width=0.9\textwidth]{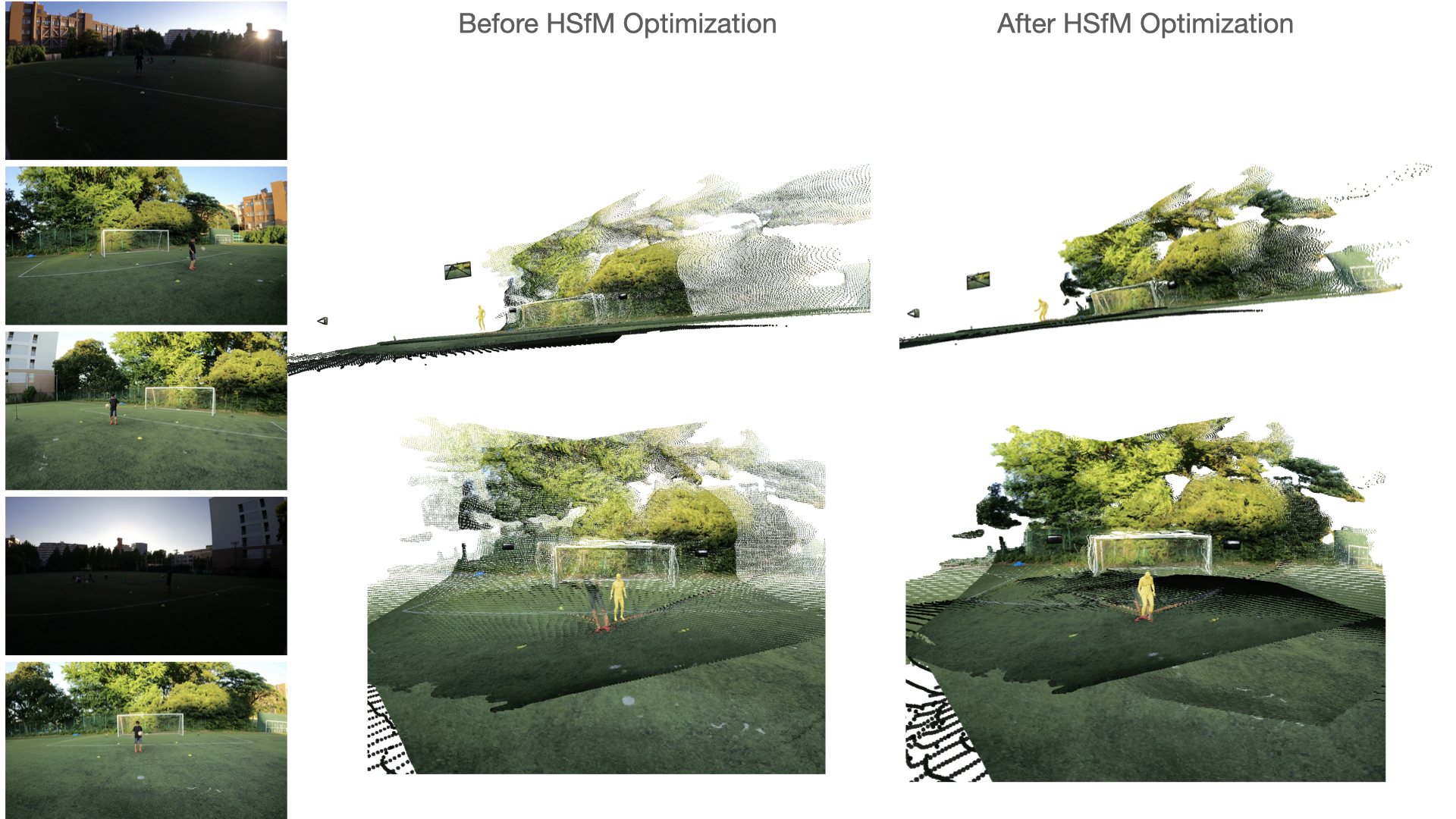}
    \end{subfigure}
    
    \vspace{1cm}
    
    \begin{subfigure}[t]{0.95\textwidth}
        \centering
        \includegraphics[width=0.9\textwidth]{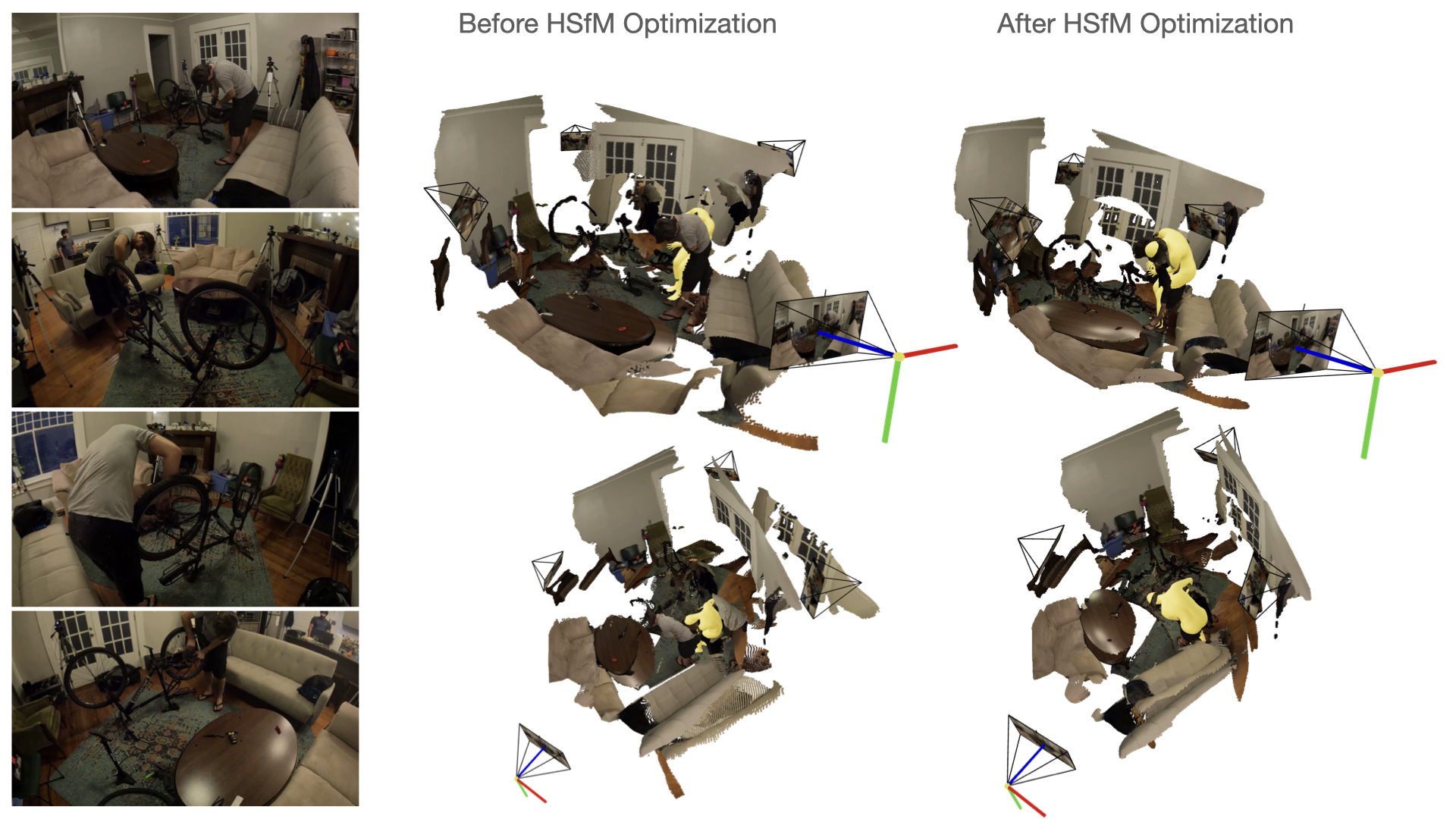}
    \end{subfigure}

    \caption{\textbf{Qualitative results.} We show reconstruction on EgoExo4D. On the left, the input images to our method, the scene, humans, and cameras before optimization (HSfM (init.)) in the center, and the reconstruction of our method after joint optimization on the right.}
    \label{fig:stacked_images_01}
\end{figure*}

\begin{figure*}[htbp]
    \centering
    \begin{subfigure}[t]{0.95\textwidth}
        \centering
        \includegraphics[width=0.9\textwidth]{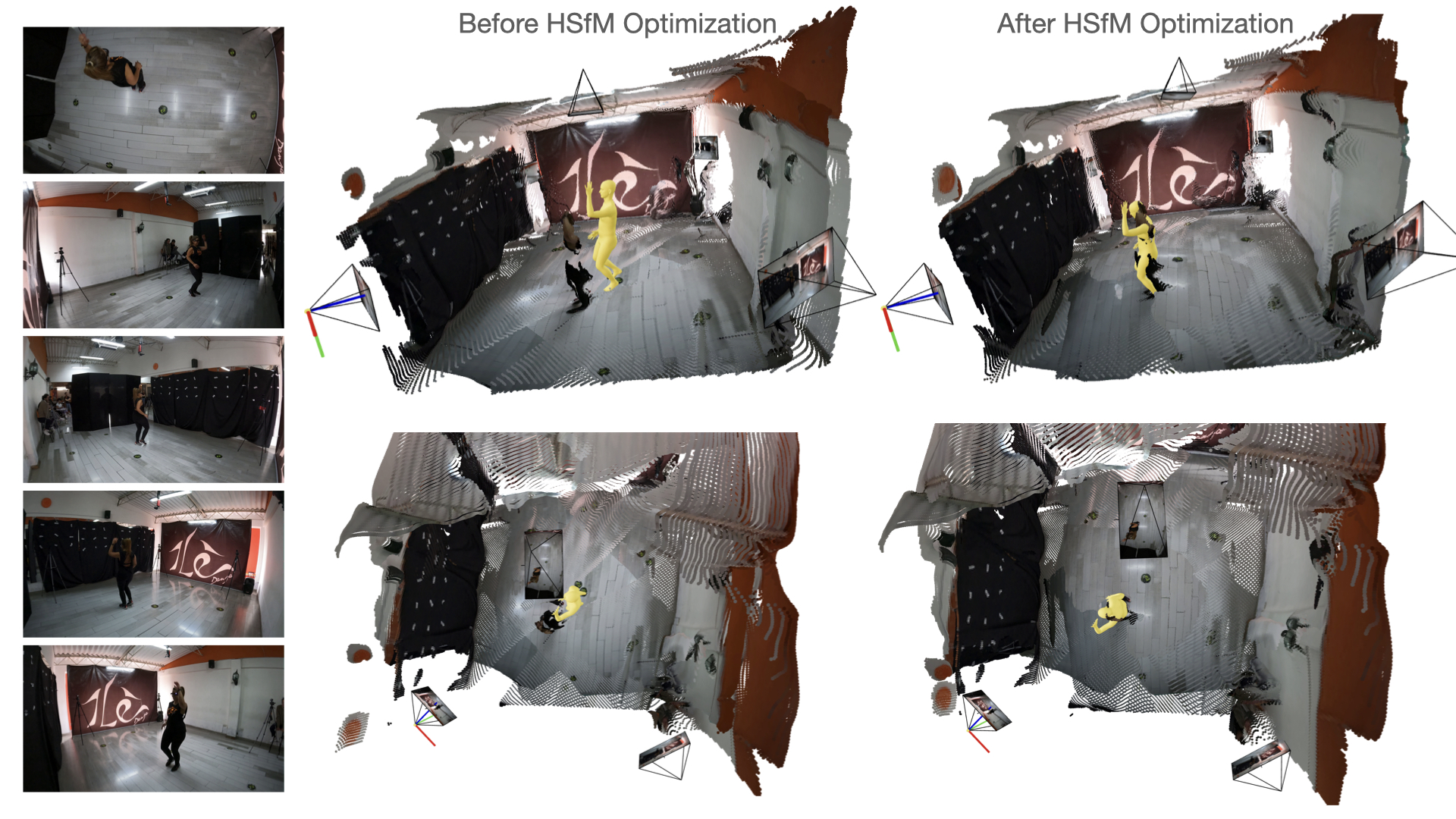}
    \end{subfigure}
    
    \vspace{1cm}
    
    \begin{subfigure}[t]{0.95\textwidth}
        \centering
        \includegraphics[width=0.9\textwidth]{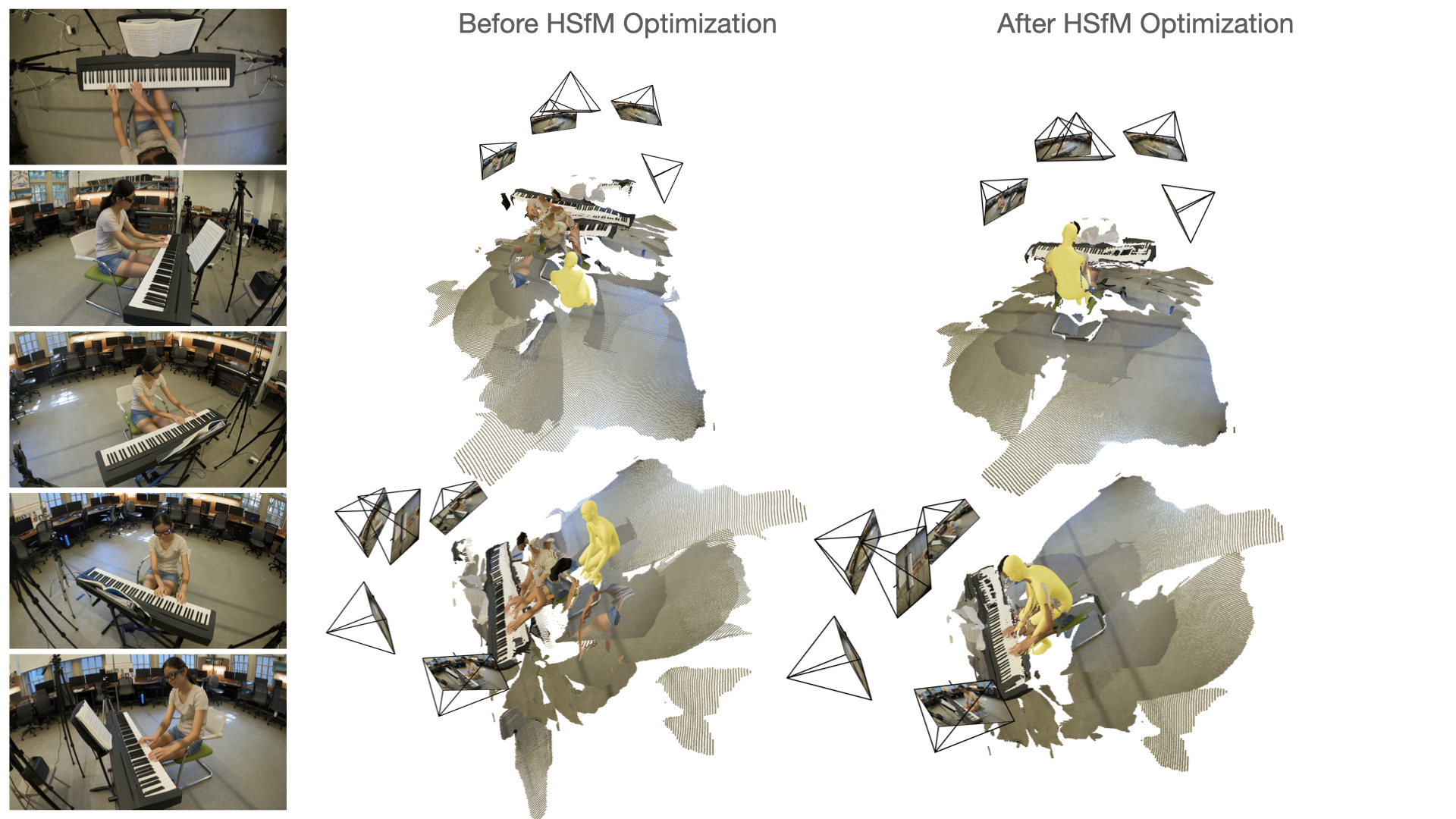}
    \end{subfigure}

    \caption{\textbf{Continuation of \cref{fig:stacked_images_01}}}
    \label{fig:stacked_images_02}
\end{figure*}

\begin{figure*}[htbp]
    \centering
    \begin{subfigure}[t]{0.95\textwidth}
        \centering
        \includegraphics[width=0.9\textwidth]{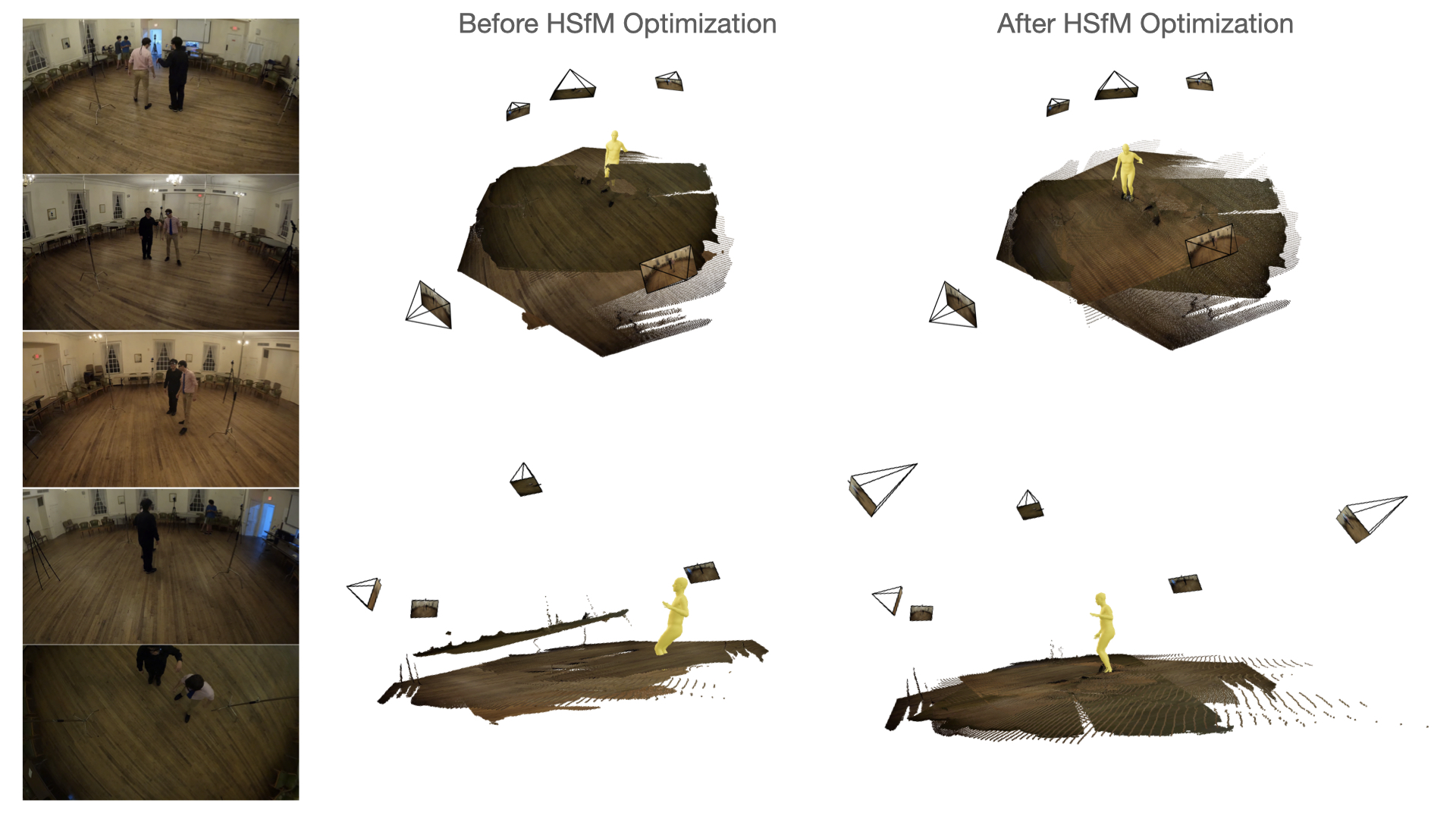}
    \end{subfigure}
    
    \vspace{1cm}
    
    \begin{subfigure}[t]{0.95\textwidth}
        \centering
        \includegraphics[width=0.9\textwidth]{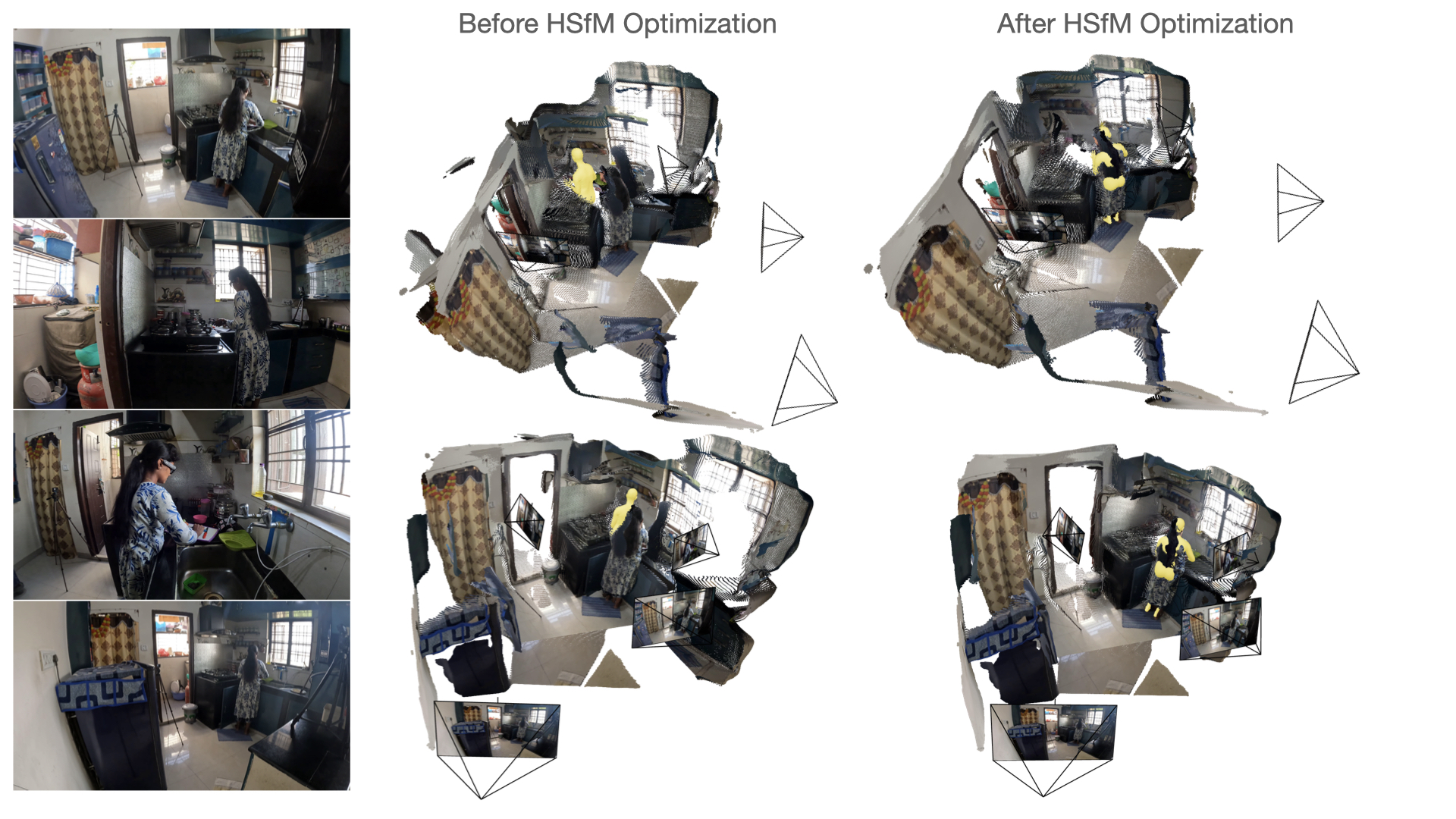}
    \end{subfigure}

    \caption{\textbf{Continuation of \cref{fig:stacked_images_01}}}
    \label{fig:stacked_images_03}
\end{figure*}

\begin{figure*}[htbp]
    \centering
    \begin{subfigure}[t]{0.95\textwidth}
        \centering
        \includegraphics[width=0.9\textwidth]{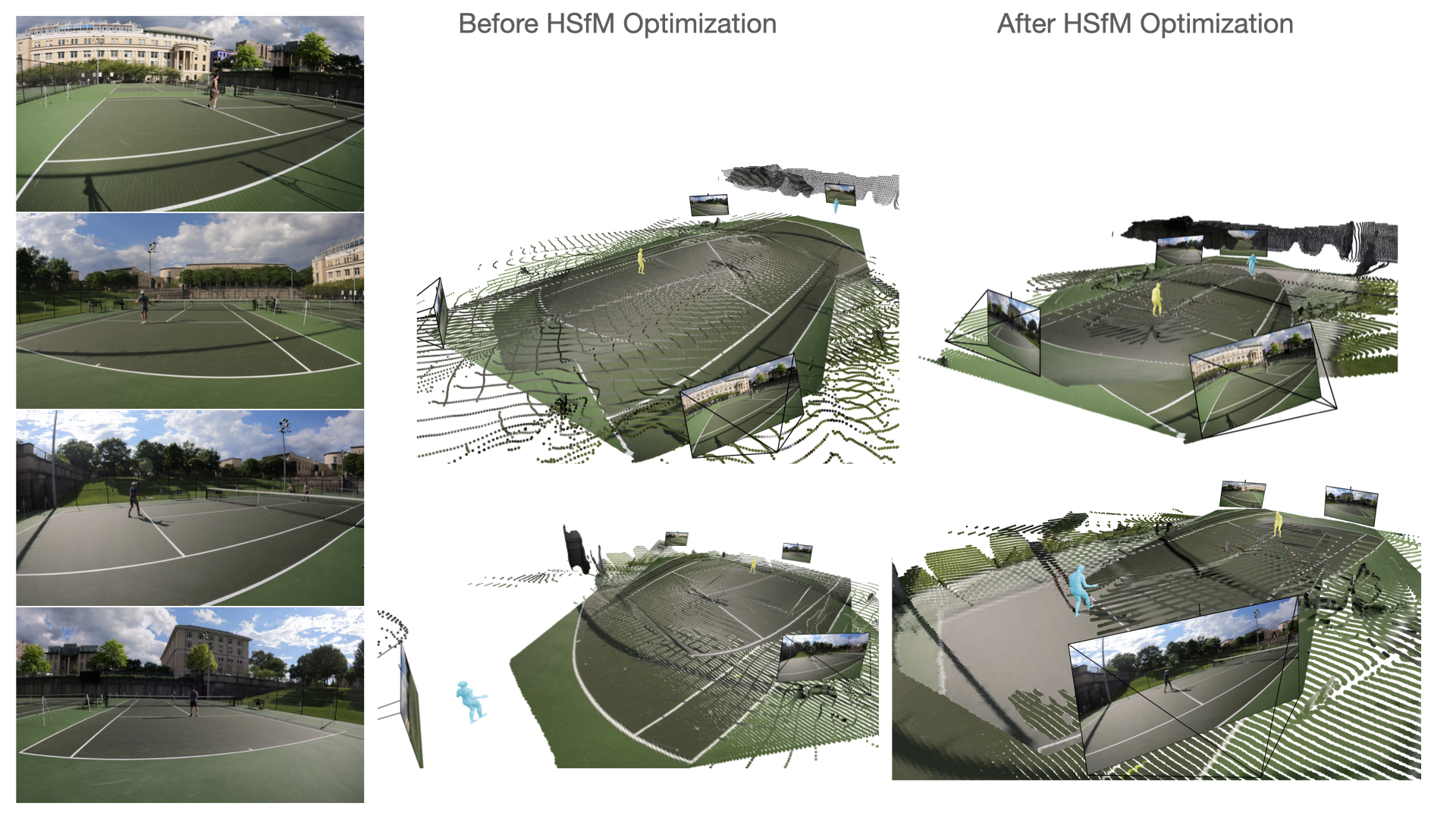}
    \end{subfigure}
    
    \vspace{1cm}
    
    \begin{subfigure}[t]{0.95\textwidth}
        \centering
        \includegraphics[width=0.9\textwidth]{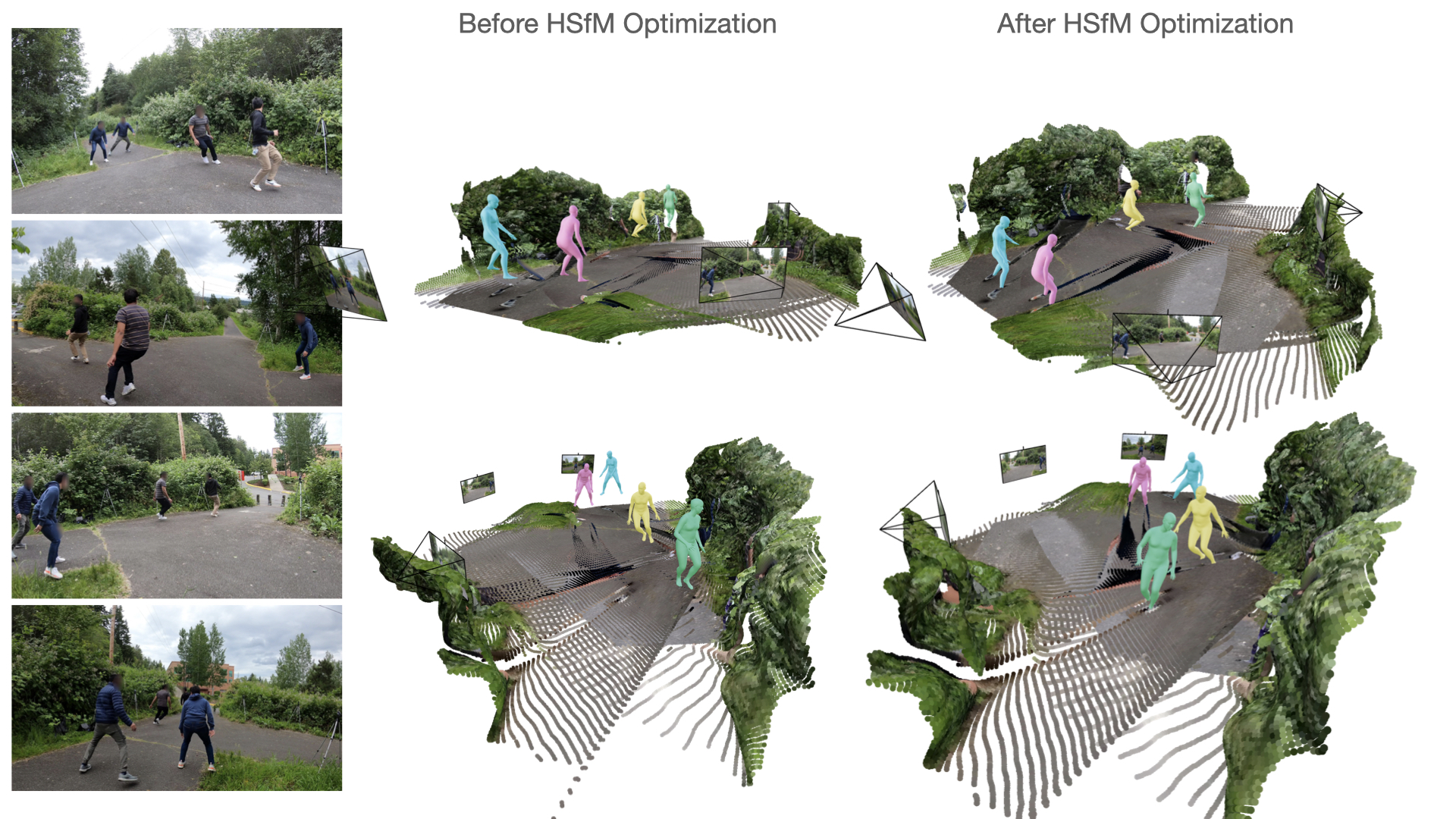}
    \end{subfigure}

    \caption{\textbf{Qualitative results.} We show reconstructions on EgoHumans. On the left, the input images to our method, the scene, humans, and cameras before optimization (HSfM (init.)) in the center, and the reconstruction of our method after joint optimization on the right. }
    \label{fig:stacked_images_04}
\end{figure*}

\begin{figure*}[htbp]
    \centering
    \begin{subfigure}[t]{0.95\textwidth}
        \centering
        \includegraphics[width=0.9\textwidth]{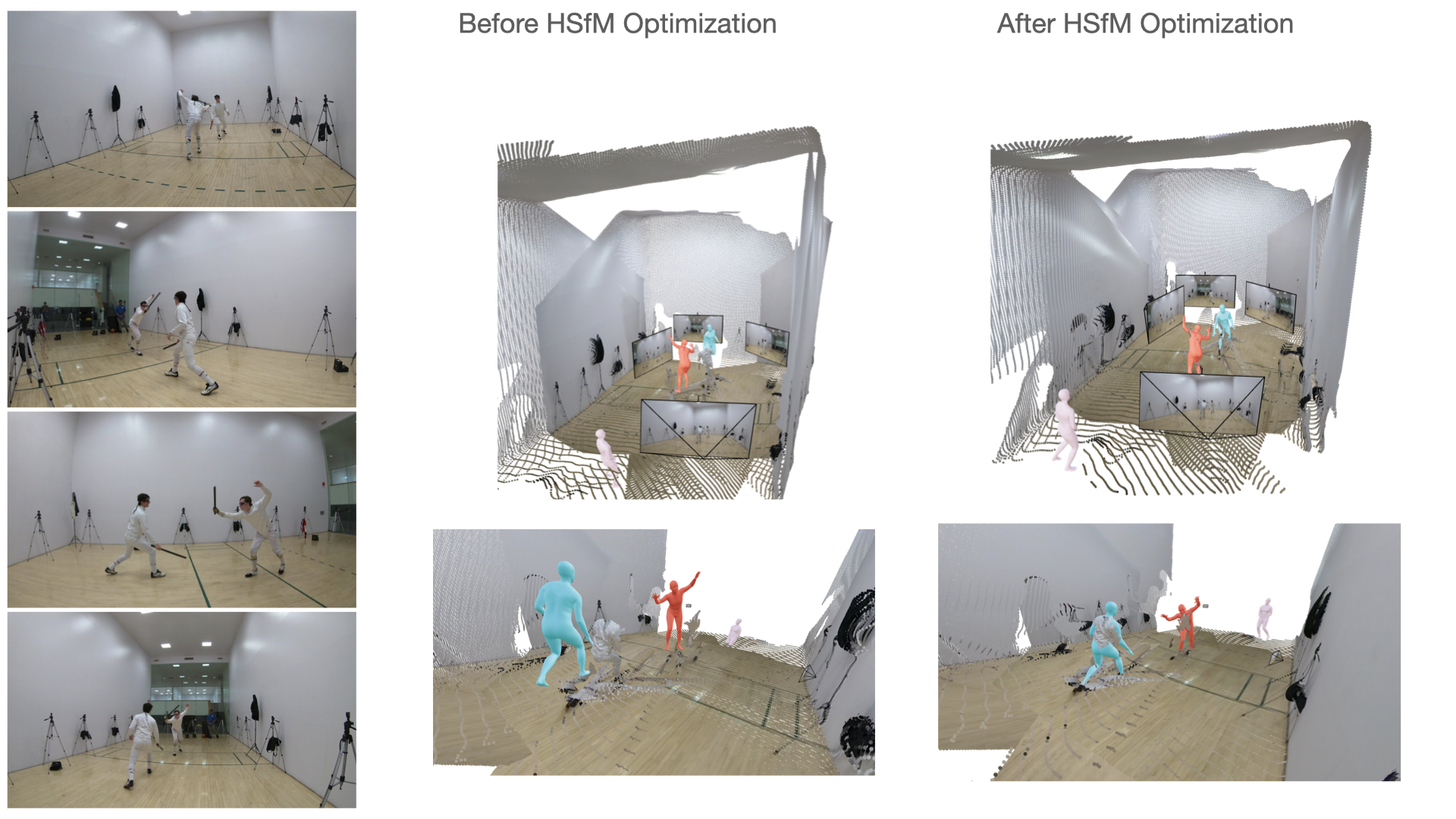}
    \end{subfigure}
    
    \vspace{1cm}
    
    \begin{subfigure}[t]{0.95\textwidth}
        \centering
        \includegraphics[width=0.9\textwidth]{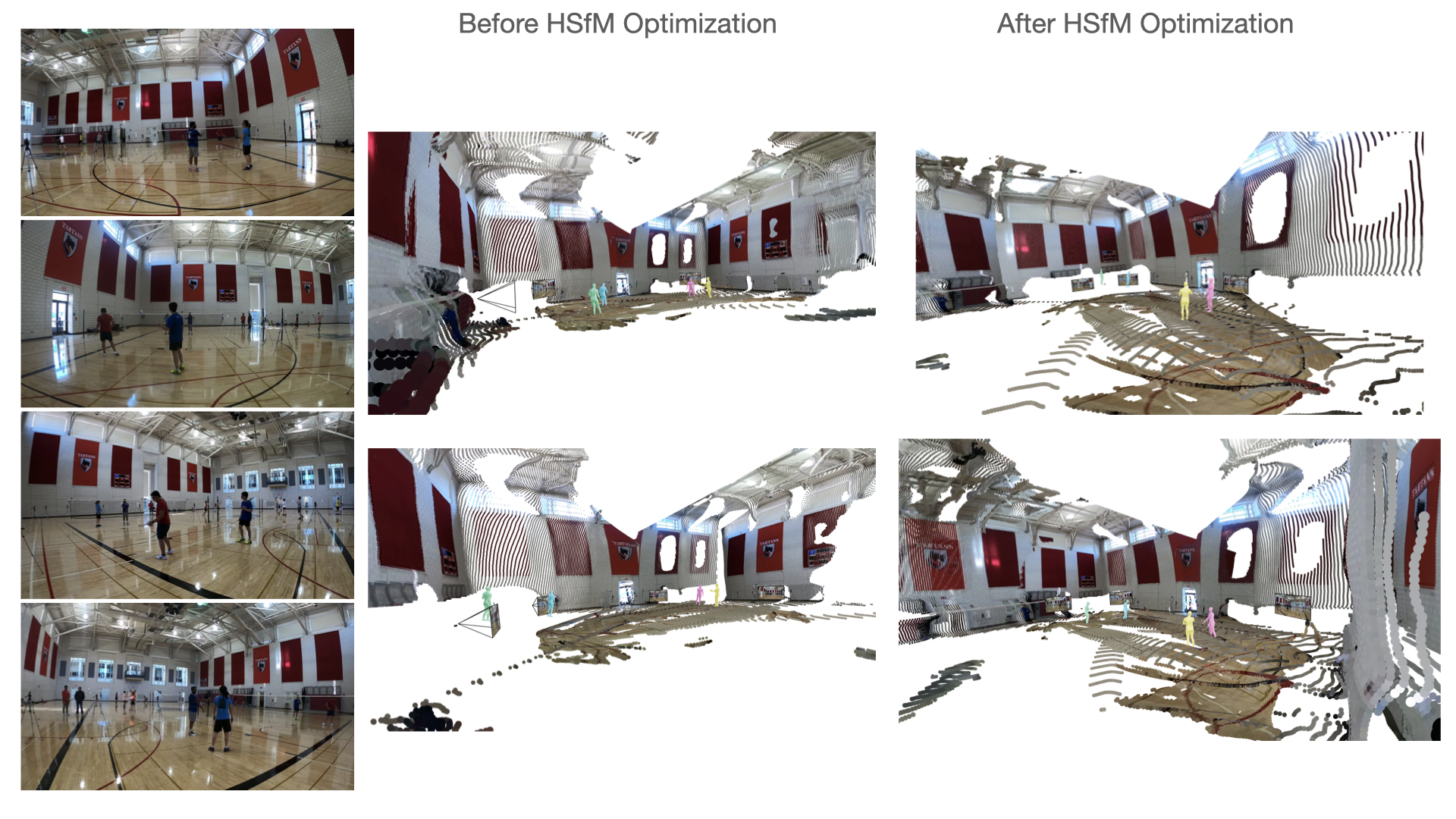}
    \end{subfigure}

    \caption{\textbf{Continuation of \cref{fig:stacked_images_04}}}
    \label{fig:stacked_images_05}
\end{figure*}




    
    